\documentclass[10pt,twocolumn,letterpaper]{article}

\usepackage{cvpr}              % To produce the CAMERA-READY version
\usepackage[accsupp]{axessibility}

\definecolor{rblue}{RGB}{68, 114, 196}
\definecolor{rgreen}{RGB}{112, 173, 71}
\definecolor{rorange}{RGB}{237, 125, 49}
\definecolor{rred}{RGB}{255, 0, 0}

\definecolor{cvprblue}{rgb}{0.21,0.49,0.74}
\usepackage[pagebackref,breaklinks,colorlinks,allcolors=cvprblue]{hyperref}
\usepackage{amssymb}
\usepackage{algorithm,algorithmicx,algpseudocode}
\usepackage{graphicx}
\usepackage{multirow}
\usepackage{pifont}
\usepackage{dsfont}
\usepackage{soul}

\title{Rethinking Diffusion for Text-Driven Human Motion Generation: Redundant Representations, Evaluation, and Masked Autoregression}

\author{
Zichong Meng,~~~Yiming Xie,~~~Xiaogang Peng,~~~Zeyu Han,~~~Huaizu Jiang \\
Northeastern University\\
\tt\small $\{$meng.zic, xie.yim, peng.xiaog, han.zeyu, h.jiang$\}$@northeastern.edu \\
{\tt\small\textbf{\href{https://neu-vi.github.io/MARDM/}{https://neu-vi.github.io/MARDM/}}}
}

\begin{document}
\maketitle
\begin{abstract}
%vq sota but due to performance metrics
Since 2023, Vector Quantization (VQ)-based discrete generation methods have rapidly dominated human motion generation, primarily surpassing diffusion-based continuous generation methods in standard performance metrics.
%vq limitations
However, VQ-based methods have inherent limitations. Representing continuous motion data as limited discrete tokens leads to inevitable information loss, reduces the diversity of generated motions, and restricts their ability to function effectively as motion priors or generation guidance.
%why diffusion back
In contrast, the continuous space generation nature of diffusion-based methods makes them well-suited to address these limitations and with even potential for model scalability.
%our study
In this work, we systematically investigate why current VQ-based methods perform well and explore the limitations of existing diffusion-based methods from the perspective of motion data representation and distribution.
%we propose
Drawing on these insights, we preserve the inherent strengths of a diffusion-based human motion generation model and gradually optimize it with inspiration from VQ-based approaches. Our approach introduces a human motion diffusion model enabled to perform masked autoregression, optimized with a reformed data representation and distribution. 
Additionally, we propose a more robust evaluation method to assess different approaches.
Extensive experiments on various datasets demonstrate our method outperforms previous methods and achieves state-of-the-art performances.
\end{abstract}    
\vspace{-2.5em}
\section{Introduction}
\label{sec:intro}
In this paper, we study the problem of human motion generation from the textual prompt (\eg, a person walks). 
Due to the remarkable performance in the image generation domain~\cite{ddpm, ldm, imagen}, diffusion models are largely adopted for human motion generation starting with the pioneer methods~\cite{mdm, motiondiffuse, flame}. 
Compared with RNN~\cite{rnn}-based generation methods~\cite{jl2p, d2m, s2s, humanml3d}, diffusion-based models offer a simpler training objective and improved stability.

In 2023, the exploration of Vector Quantization (VQ) techniques for human motion representation becomes increasingly dominant,
marked a noticeable shift in attention away from diffusion models for the human motion generation task~\cite{tm2t, t2m-gpt, momask}.
These methods transform continuous motion representations (\eg processed joint positions) into discrete tokens, which enables the use of already proven generative architectures~\cite{transformer, bert, gpt} and their training and sampling techniques from the field of natural language processing with minimal modifications.

\begin{figure}[tb]
\centering
\includegraphics[width=0.79\linewidth, height=4.5cm]{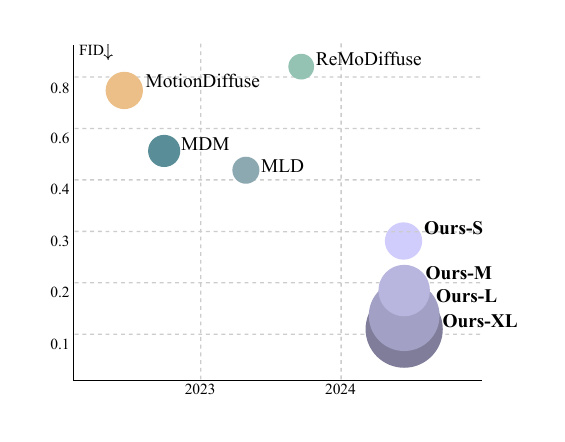}
\vspace{-2.25em}
\caption{\textbf{The FID results on HumanML3D dataset.} The bubble size is proportional to the model size. We achieve superior performance and demonstrate model scalability.
% with a diffusion-based approach.
}
\label{fig:model_scale}
\vspace{-2em}
\end{figure}

Despite the performance improvement, VQ-based methods still exhibit notable limitations. Representing continuous motion data as limited groups of discrete tokens inherently causes a loss of motion information, reduces generation diversity, and limits their ability to serve as motion priors or generation guidance (\eg, in the generation of dual-human motion and human-object interactions). 
Moreover, unlike language models, these discrete tokens often lack contextual richness, which can hinder model scalability.

In contrast, the continuous space nature of diffusion-based generation methods effectively overcomes these limitations and offers potential for model scalability, as evidenced by numerous image diffusion models~\cite{dit, mar}. 
In light of these, researchers have started to revisit diffusion-based approaches~\cite{motionmamba, remodiffuse}. 
However, these attempts struggle to achieve performance comparable to that of VQ-based methods.
More importantly, the reasons for the performance gap between VQ and diffusion-based human motion generation methods remain unclear.

In this work, we first systematically investigate why VQ-based motion generation approaches perform well and explore the limitations of diffusion-based methods from the perspective of motion data representation and distribution. Specifically, 
(1) We examine how VQ-based discrete formulations benefit from training with current motion data representation, which consists of redundant dimensions, while this data composition and its distribution hinder existing diffusion models.
(2) We explore how VQ-based methods inherently align with current evaluation metrics, which incorporate the entire data representation including redundant dimensions, whereas diffusion-based methods are often penalized under these evaluation criteria.

Based on our diagnostic findings and inspirations from current VQ-based approaches, we aim to close the performance gap by gradually enhancing a diffusion-based model tailored for human motion generation.
We first restructure the motion data representation by removing redundant information, resulting in a distribution better suited for diffusion models. We then process them with a 1D ResNet~\cite{resnet}-based AutoEncoder for better per-frame smoothness.
Additionally, we build a diffusion-based motion generation model with masked autoregressive strategies to simplify training objective.
Finally, we propose more robust evaluators for unbiased assessments of different approaches.

We summarize our contributions as follows:
 \begin{itemize}
    \item We systematically investigate the reasons why VQ-based methods outperform diffusion-based methods from motion data representation and distribution perspective, providing analysis with theoretical and experimental support.
    \item Inspired by our diagnostic findings, we propose a scalable masked autoregressive diffusion-based generation framework and a more robust evaluation method.
    \item Our method achieves state-of-the-art performance on text-to-motion generation task with significant improvements on KIT-ML~\cite{kit} and HumanML3D~\cite{humanml3d} datasets.
 \end{itemize}
\section{Diagnosis: Motion Data Representation and Distribution}
\label{sec:diagnosis}
In this section, we study how motion representation and distribution impact training, sampling (\cref{sec:diagnosis1}), and evaluation robustness (\cref{sec:diagnosis2}), revealing how these factors favor VQ-based methods but limit diffusion-based approaches.

\subsection{Prelimenary}
\label{sec:diagnosis_pre}
\noindent\textbf{VQ-based Human Motion Generation Methods} primarily adopt either a standard Vector Quantized Variational AutoEncoders (VQ-VAEs)~\cite{vqvae, t2m-gpt} or a residual (R) VQ-VAE~\cite{momask} to map motion data into discrete tokens.

Given a motion sequence $\mathbf{x}^{1:N
} \in \mathbb{R}^{N \times D}$ of length $N$ and human pose dimension $D$, the transformation begins by encoding $\mathbf{x}^{1:N}$ into a latent sequence $\mathbf{h}^{1:n
} \in \mathbb{R}^{n \times d}$ with a 1D convolutional encoder $\boldsymbol{\mathrm{E}}$.
For vanilla VQ-VAE, each vector is quantized via a base VQ codebook to the nearest token.
In RVQ-VAE, additional residual quantization layers are used to quantify the difference between the original latent vector and the quantized representation from the preceding layers.
The indices $\mathbf{k}$ from the base layer (VQ-VAE) or all layers (RVQ-VAE) form discrete inputs for training generative models with sequence generation.

The generated discrete sequence $\mathbf{g} = \mathbf{g}_{0:n}$ (or $\mathbf{g}_{0:n}^{0:V+1}$ in residual case) is embedded via the codebook, then projected back (or summed then project in residual case) with an upsample convolutional decoder $\boldsymbol{\mathrm{D}}$ to obtain the final motion.

\noindent\textbf{Diffusion-based Human Motion Generation Methods.}
The diffusion-based methods use an interpolation function, $\mathbf{x}_t = \alpha \mathbf{x}_0 + \sigma \boldsymbol{\epsilon}$, to combine ground truth (GT) motion data with Gaussian noise $\boldsymbol{\epsilon}$ for a noisy motion $\mathbf{x}_t$. In motion generation, typically following DDPM~\cite{ddpm}, this function is defined as:
\vspace{-1.5mm}
\begin{equation}
\label{equation1}
\mathbf{x}_t=\sqrt{\bar\alpha_t} \mathbf{x}_0+\sqrt{1-\bar\alpha_t} \boldsymbol{\epsilon}
\vspace{-1.75mm}
\end{equation}
$\bar\alpha_t$ controls the pace of the diffusion process where $0 = \bar\alpha_T < \cdots < \bar\alpha_0 = 1$ with assumption that $\mathbf{x}_T \sim \mathcal{N}(\mathbf{0}, \mathbf{I})$.

During training, the model learns to predict a continuous vector from $\mathbf{x}_t$ given $t$, usually the noise $\boldsymbol{\epsilon}$ or original motion $\mathbf{x}_0$, and is optimized with a mean squared error (MSE) loss between the predicted value and its ground truth.

During sampling, starting from random noise $\mathbf{x}_T \sim \mathcal{N}(\mathbf{0}, \mathbf{I})$, for each $t$, the model with $\mathbf{x}_{t}$ as input, predicts the original motion $\mathbf{x}_0$ or the noise $\boldsymbol{\epsilon}$ (then to $\mathbf{x}_0$ with \cref{equation1}), and deduces intermediate samples $\mathbf{x}_{t-1}$ (for $t \in {1 \text{ to } T}$):
\vspace{-2mm}
\begin{equation}
    \frac{\sqrt{\alpha_t}(1 - \bar{\alpha}_{t-1})}{1 - \bar{\alpha}_t} \mathbf{x}_t + \frac{\sqrt{\bar{\alpha}_{t-1}}(1 - \alpha_t)}{1 - \bar{\alpha}_t} \mathbf{x}_0 + \sqrt{1 - \bar{\alpha}_t} \boldsymbol{\epsilon}_t
    \vspace{-2mm}
\end{equation}
where $\boldsymbol{\epsilon}_t\sim \mathcal{N}(0, \mathbf{I})$, until it reaches the clean motion $\mathbf{x}_0$.

\noindent\textbf{Motion Data Representation.}
The majority of recent methods utilize the canonical pose representation introduced by \cite{humanml3d} on widely-used datasets, including KIT-ML~\cite{kit} and HumanML3D~\cite{humanml3d}. This representation at a given time step $i$ is defined as $\mathbf{x}^{i} = [\dot{r}^{a}, \dot{r}^{xz}, \dot{r}^{h}, j^p, j^v, j^r, c^f]$, comprising seven feature components: root angular velocity $\dot{r}^{a}$, root linear velocities $\dot{r}^{xz}$ in the XZ-plane, root height $\dot{r}^{h}$, local joint positions $j^p \in \mathbb{R}^{3(N_j-1)}$, local velocities $j^v \in \mathbb{R}^{3(N_j-1)}$, joint rotations $j^r \in \mathbb{R}^{6(N_j-1)}$ in local space, and binary foot-ground contact features $c^f \in \mathbb{R}^{4}$, where $N_j$ denotes the joint number. However, only the first 4 feature groups from this over-parameterized representation are used to produce final human motion, making the remaining 3 components redundant.
We categorize the first 4 feature groups as essential features while classifying the remaining as redundant.

\subsection{Impacts on Training and Sampling}
\label{sec:diagnosis1}
\begin{table}[t]
\renewcommand{\arraystretch}{1.0}
\caption{\textbf{Impact of redundant features on VQ-based models.} VQ-based methods, T2M-GPT and MoMask, trained with redundant features exhibit better reconstruction performance and lead to better generation quality on the HumanML3D dataset.}
\vspace{-2em}
\label{tab:diagnosis1} 
\begin{center}
\resizebox{0.85\linewidth}{!}{\begin{tabular}{c|c|c|c|c|c|c}
\toprule
\multirow{2}{*}{{Method}} & Trained With & \multicolumn{2}{c|}{FID $\downarrow$} & \multicolumn{3}{c}{R-Precision $\uparrow$} \\
 \cmidrule(r){3-7}
 & Redundancy & Recon  & Gen  &Top 1 &Top 2 & Top 3 \\
\midrule
     T2M-GPT~\cite{t2m-gpt}  & \ding{51} & $0.081$ &$0.335$& $0.470$ &$0.659$  & $0.758$ \\
    \midrule
      T2M-GPT~\cite{t2m-gpt} & \ding{55} & $0.095$ & $0.418$ &$0.466$ &$0.653$  & $0.753$\\
    \midrule
     MoMask~\cite{momask} & \ding{51} &$0.029$ & $0.116$ &$0.490$ & $0.687$  &$0.786$ \\
    \midrule
     MoMask~\cite{momask} & \ding{55} & $0.030$ & $0.200$ & $0.485$& $0.681$  & $0.782$ \\
\bottomrule
\end{tabular}}
\vspace{-1mm}
\end{center}
\end{table}
\begin{figure}[tb]
\vspace{-1.5em}
  \centering
  \begin{subfigure}{\linewidth}
    \centering
    \begin{subfigure}{0.495\linewidth}
      \includegraphics[width=\linewidth]{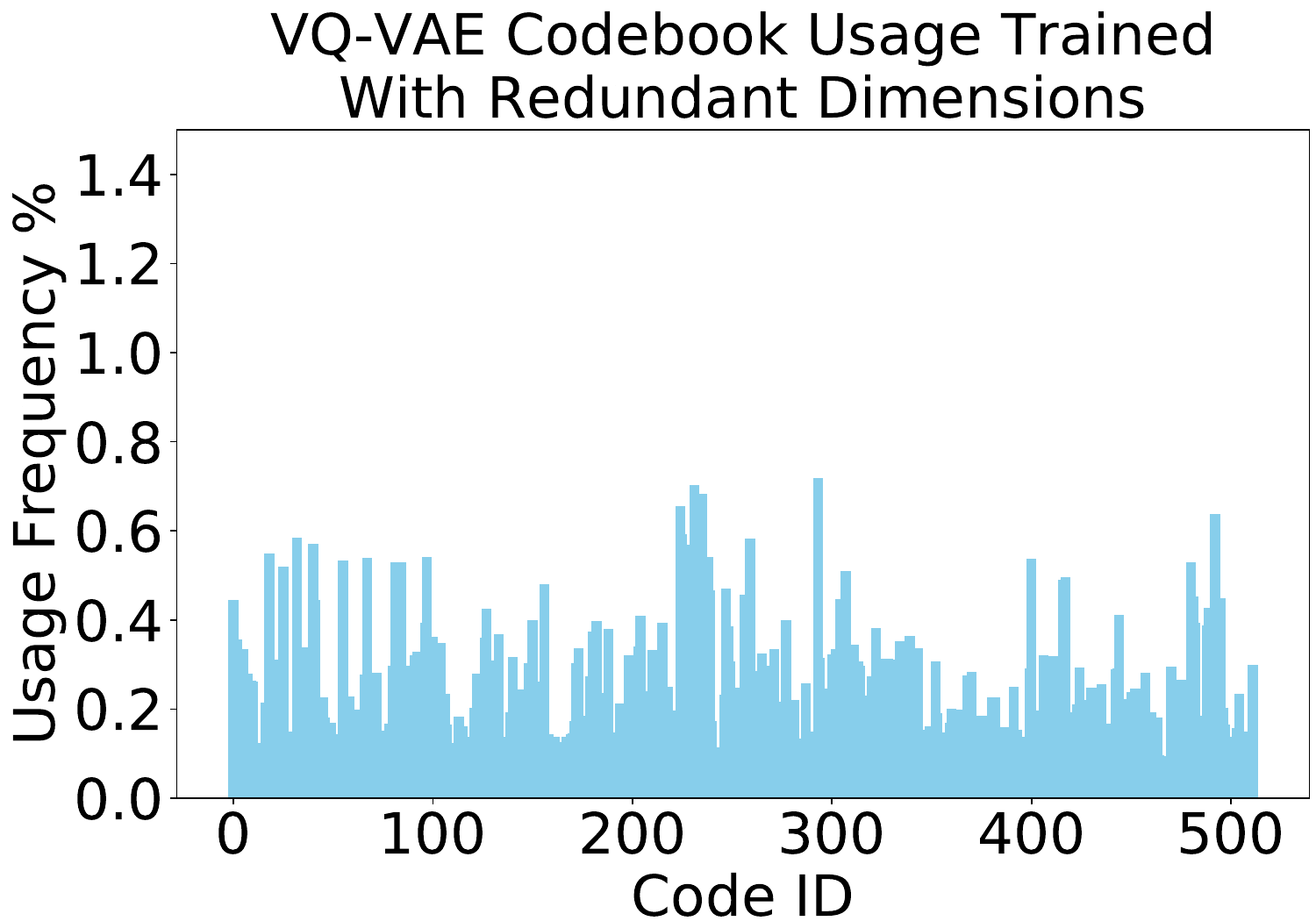}
    \end{subfigure}
    \begin{subfigure}{0.495\linewidth}
      \includegraphics[width=\linewidth]{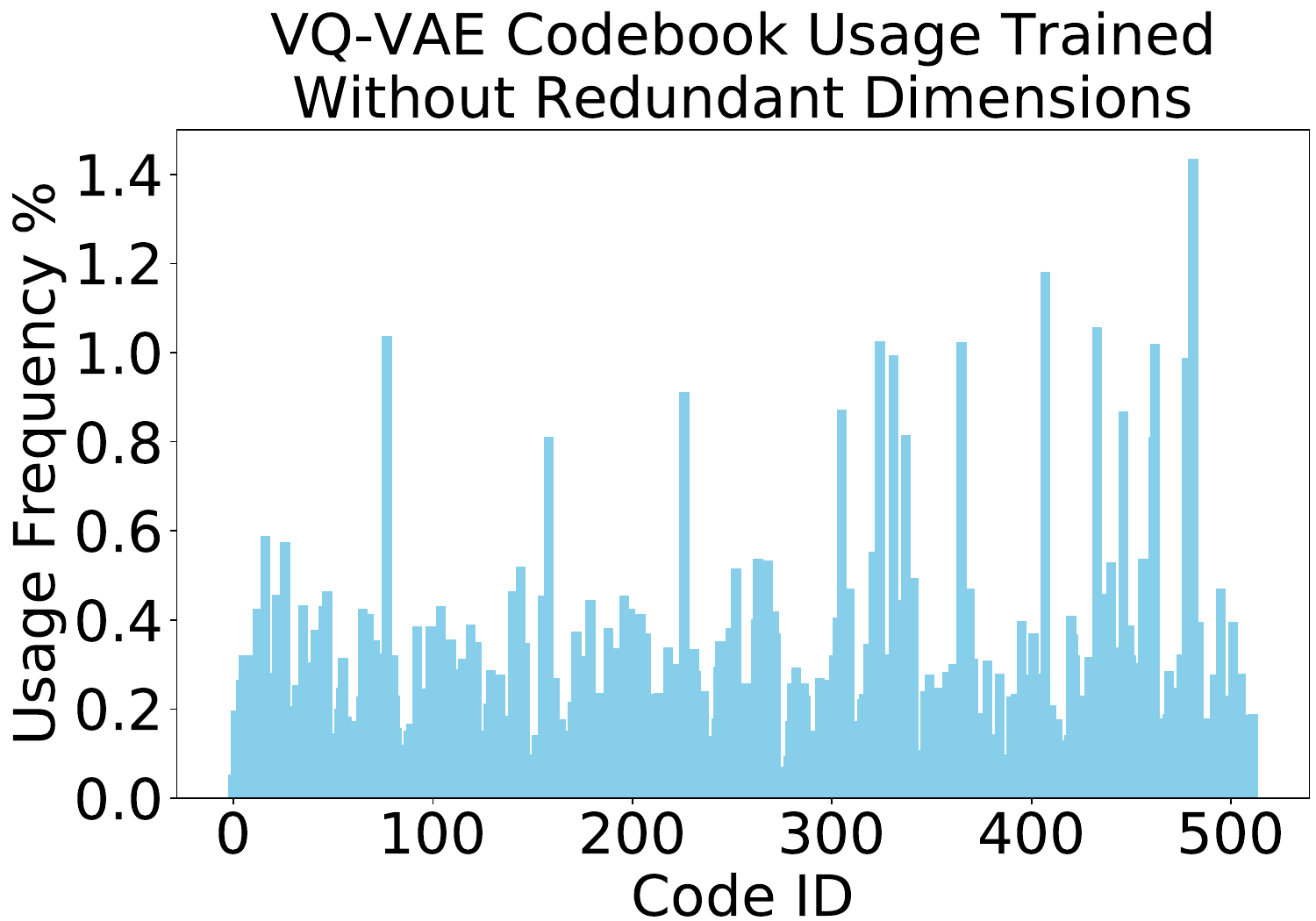}
    \end{subfigure}
  \end{subfigure}
    \begin{subfigure}{\linewidth}
    \centering
    \begin{subfigure}{0.49\linewidth}
      \includegraphics[width=\linewidth]{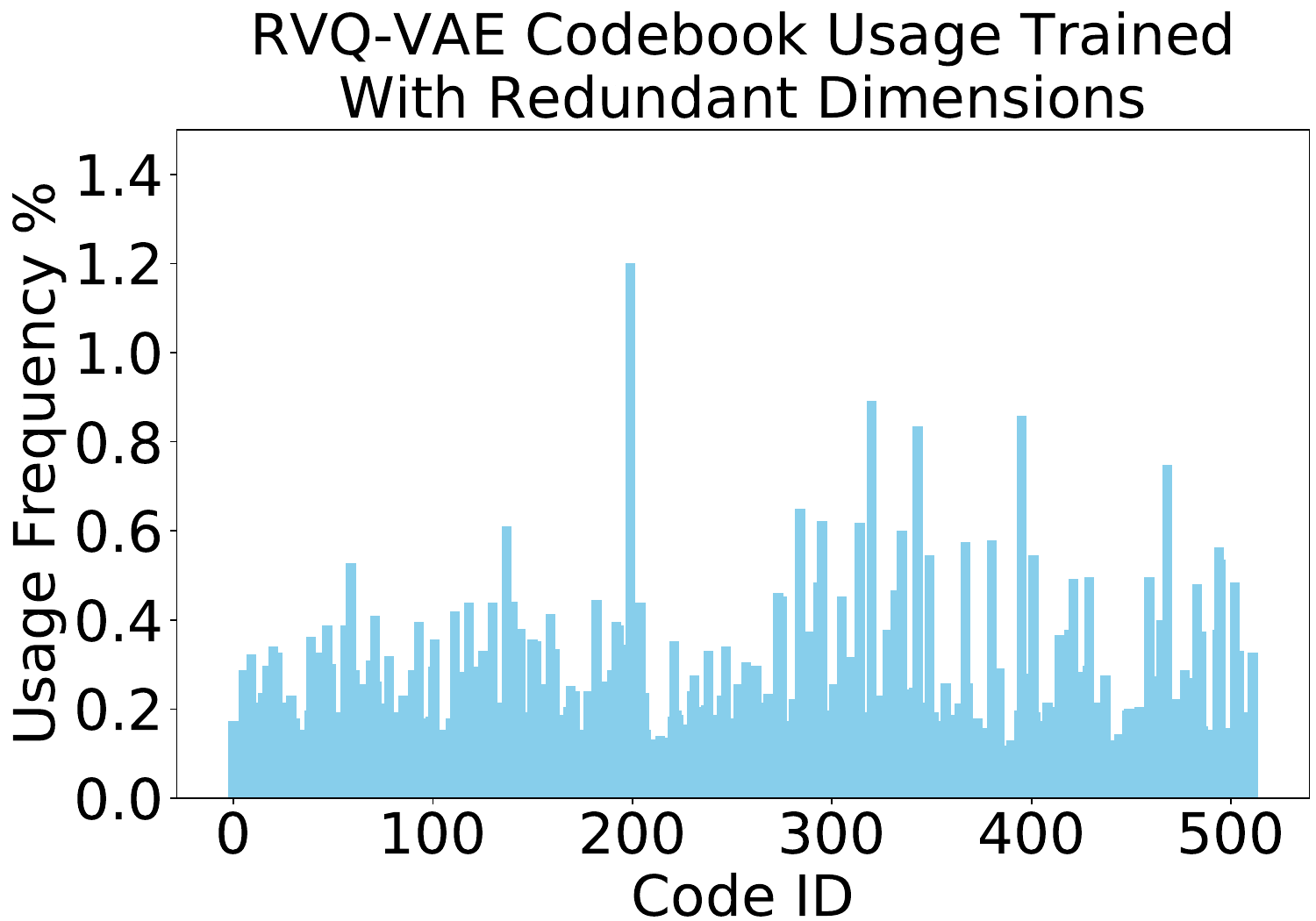}
    \end{subfigure}
    \begin{subfigure}{0.49\linewidth}
      \includegraphics[width=\linewidth]{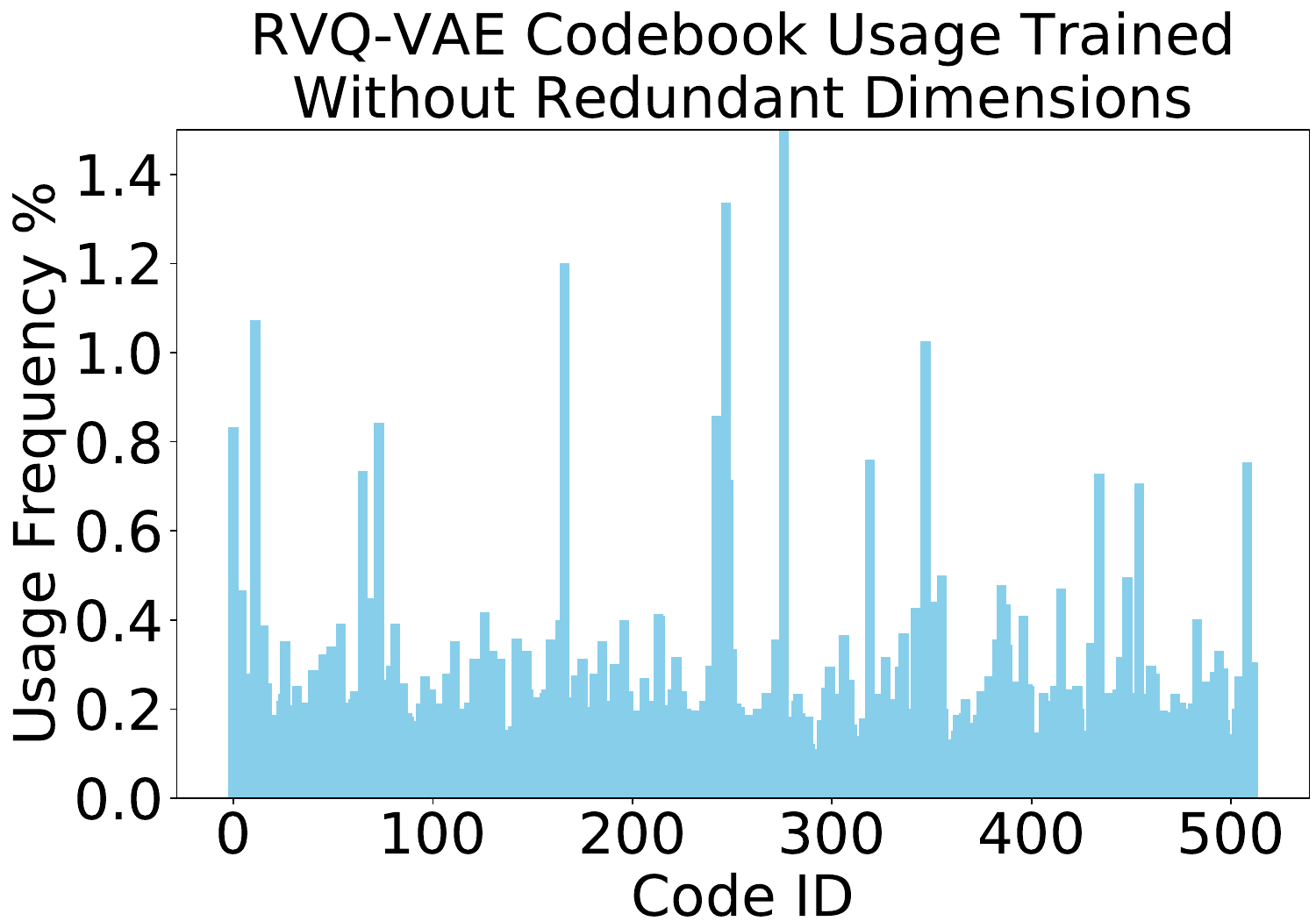}
    \end{subfigure}
  \end{subfigure}
  \vspace{-1.5em}
  \caption{\textbf{Code Usage of VQ-VAEs} trained with redundancy are more balanced than VQ-VAEs trained with only essential features.}
  \label{fig:diagnosis1}
\vspace{-1.5em}
\end{figure}
In this section, we present how this motion representation and distribution impact VQ and diffusion-based methods.

\noindent\textbf{Benefiting VQ-Based Methods.}
The redundancy in data representation benefits VQ-VAE training which then enhances discrete generative modeling. 
To validate this viewpoint, we conduct ablative experiments by training VQ-VAE from T2M-GPT~\cite{t2m-gpt} and RVQ-VAE from MoMask~\cite{momask} on HumanML3D, comparing their reconstruction performance with and without redundant data representations.
Then we train T2M-GPT and MoMask following their original methods, both with and without redundant data representations, and evaluate their performance using human motion generation metrics. 
As shown in~\cref{tab:diagnosis1}, training with redundant dimensions results in VQ-VAEs with significantly better reconstruction performance, which subsequently enhances the performance of discrete generative models. 
To understand the reason, we further analyze the role of data representation in VQ-VAE training.

Let $\mathbf{x}^{\text{GT}}_{\text{r}}$ and $\mathbf{x}^{\text{GT}}_{\text{e}}$ denote the ground truth data with and without redundancy and $\mathbf{x}^{\text{pred}}_{\text{r}}$ and $\mathbf{x}^{\text{pred}}_{\text{e}}$ represent the prediction. The reconstruction loss $\mathcal{L}^{\text{rec}}_r$ with redundancy can be decomposed as:
\vspace{-2mm}
\begin{align}
\label{equation3}
    L^{\text{rec}}_{r} &= \mathcal{L}(\mathbf{x}^{\text{GT}}_{\text{e}} - \mathbf{x}^{\text{pred}}_{\text{e}} +\mathbf{x}^{\text{GT}}_{\text{r-e}} -\mathbf{x}^{\text{pred}}_{\text{r-e}}) \\
    &=L^{\text{rec}}_{e} + \mathcal{L}(\mathbf{x}^{\text{GT}}_{\text{r-e}} -\mathbf{x}^{\text{pred}}_{\text{r-e}}). \notag
    \vspace{-2mm}
\end{align}
where $\mathcal{L}^{\text{rec}}_e$ is the reconstruction loss without redundancy representation, and $\mathcal{L}$ is typically measured with variation of $L1$ or $L2$ loss.
This decomposition shows the redundancy acts as a data-level regularization

This built-in regularization reduces model variance, making representations less sensitive to training data fluctuations. 
This improvement enhances generalization, ultimately leading to more effective and consistent codebook usage,
as shown in \cref{fig:diagnosis1}. 
Specifically, on the HumanML3D test set, both VQ-VAE and RVQ-VAE, which were trained on the data with redundant dimensions, exhibit more uniform code utilization. In contrast, models trained on the data without redundancy show distinct spikes in certain codes. Numerically, for VQ-VAE, the KL divergence against uniform distribution is 0.205 \vs 0.337, and for RVQ-VAE, it is 0.243 \vs 0.351, using original \vs essential dimensions.
These results demonstrate the benefit of incorporating redundancy for VQ-based methods.

\begin{table}[t]
\renewcommand{\arraystretch}{1}
\caption{\textbf{The results of MDM on humanML3D dataset}. We report the results of MDM with original $\mathbf{x}_0$ prediction \vs with $\boldsymbol{\epsilon}$ prediction. Training to predict $\mathbf{x}_0$ leads to better results.}
\vspace{-2em}
\label{tab:diagnosis2} 
\begin{center}
\resizebox{0.8\linewidth}{!}{\begin{tabular}{c|c|c|c|c|c}
\toprule
\multirow{2}{*}{{Method}} & \multirow{2}{*}{{Prediction}}& \multicolumn{1}{c|}{FID $\downarrow$} & \multicolumn{3}{c}{R-Precision $\uparrow$} \\
 \cmidrule(r){3-6}
 & & Gen  &Top 1 &Top 2 & Top 3 \\
\midrule
    MDM-50Step~\cite{mdm} & $\mathbf{x}_0$ & $0.518$ &$0.440$ & $0.636$& $0.742$ \\
    \midrule
    MDM-50Step~\cite{mdm}-Cosine & $\boldsymbol{\epsilon}$ & $31.265$ &$0.054$ &$0.103$  & $0.147$\\
    \midrule
    MDM-50Step~\cite{mdm}-Linear & $\boldsymbol{\epsilon}$ & $1.574$ & $0.279$ & $0.336$ & $0.415$\\
\bottomrule
\end{tabular}}
\vspace{-2.5em}
\end{center}
\end{table}
\noindent\textbf{Limiting Diffusion-Based Methods.}  
The current data representation and distribution impose constraints on the modeling approaches for diffusion-based methods.

Most diffusion-based methods follow the DDPM \cite{ddpm} but predict original motion $\mathbf{x}_{0}$. However, this deterministic $\mathbf{x}_0$-only prediction in each timestep can limit training simplicity and sampling diversity.
Attempts to train these methods to predict noise often result in inaccurate motion, human shape \eg MDM~\cite{mdm} with cosine beta schedule, or persistent shaking \eg MDM with linear beta schedule in \cref{tab:diagnosis2}.
Below, we explain the two major factors in detail: dimensional distribution mismatch and error amplification.

First, the current motion data do not follow a standard normal distribution with standard z-normalization due to its mixed structure consisting of features from 3D continuous (\eg joint position), 6D rotatory (\eg joint rotation), and categorical (\eg foot contacts) distribution. 
This leads to a dimensional distribution mismatch and challenges the interpolation function (\cref{equation1} for DDPM).
In the forward diffusion process, due to the varying initial distributions across different feature groups of $\mathbf{x}_0$, feature groups of $\mathbf{x}_T$ may converge to their own distinct distribution by time step $T$, rather than all converging to a same standard distribution.
Consequently, in reverse diffusion process, starting from a standard distribution leads to errors in motion generation.

Second, when normalizing the human motion data, the standard deviation (SD) is averaged in each of 7 feature groups as $\sigma^{'\mathbf{x}} = \frac{\sum_{i=0}^{D-1} \sigma^{\mathbf{x}}_i}{D}$, where D is the number of dimensions in each group.
We define the ratio $\phi'^{\mathbf{x}} = \frac{\sigma^{\mathbf{x}\prime}}{\sigma^{\mathbf{x}}}$, where $\sigma^{\mathbf{x}}$ is the original SD.
Then this ratio is further adjusted by a feature bias term $\gamma$: $\phi'^{\mathbf{x}} = \gamma \times \frac{\sigma^{\mathbf{x}\prime}}{\sigma^{\mathbf{x}}}$ when the data is fed in the network.
This SD ratio will cause error amplification when predicting noise.

Suppose $\boldsymbol{\epsilon}_\theta(\mathbf{x}_t, t)$ denote the predicted noise $\boldsymbol{\epsilon}$ at time $t$ and define the squared error $\delta_{\mathbf{x}_0}$ between ground truth and predicted $\mathbf{x}_0$, and error $\delta_{\boldsymbol{\epsilon}}$ for noise prediction. Then we have $\delta_{\boldsymbol{\epsilon}} = \| \boldsymbol{\epsilon}_\theta(\mathbf{x}_t, t) - \boldsymbol{\epsilon} \|_2^2$ and
$\delta_{\mathbf{x}_0} = \left\| \frac{1}{\sqrt{\bar{\alpha}_t}} \left(\mathbf{x}_t - \sqrt{1 - \bar{\alpha}_t} \, \boldsymbol{\epsilon}_\theta(\mathbf{x}_t, t)\right) - \mathbf{x}_0 \right\|_2^2$.
Substitute $\mathbf{x}_0$ from \cref{equation1} (detailed deduction in App.~\ref{app:proof1}), we get:
\vspace{-1.5mm}
\begin{equation}
\label{equation4}
    \delta_{\mathbf{x}_0} = \left\|\frac{\sqrt{1 - \bar{\alpha}_t}}{\sqrt{\bar{\alpha}_t}}\right\|^2_2 \delta_{\boldsymbol{\epsilon}},
    \vspace{-1.5mm}
\end{equation}
an standard error relation $\delta_{\boldsymbol{\epsilon}} \to \delta_{\mathbf{x}_0}$.
If $\mathbf{x}_0$ is processed correctly, the relation between $\delta_{\mathbf{x}_0}$ and $\delta_{\boldsymbol{\epsilon}}$ only responds to time coefficient $\bar{\alpha}$.
Since the SD ratio $\phi'^{\mathbf{x}}$ applies to both predicted and ground truth $\mathbf{x}_0$, then \cref{equation4} updates to:
\vspace{-1.75mm}
\begin{align}
    \delta_{\mathbf{x}_0} \times \phi'_{i} &= \left\| \frac{\sqrt{1 - \bar{\alpha}_t}}{\sqrt{\bar{\alpha}_t}} \right\|_2^2 \delta_{\boldsymbol{\epsilon}},
    \vspace{-2.25mm}
\end{align}
where $\bar{\alpha}$ remains unchanged and $\boldsymbol{\epsilon}$ guaranteed from normal distribution.
This means unlike direct $\mathbf{x}_0$ prediction, errors from predicting $\boldsymbol{\epsilon}$ are actually amplified because of the modified SD and are worsened by feature bias.

Both dimensional distribution mismatch and noise prediction error amplification can seriously impact generation. Therefore, reformatting motion representation and distribution is crucial to improve motion diffusion modeling.

\subsection{Impact on Method Evaluation Robustness}
\label{sec:diagnosis2}
The widely adopted evaluators~\cite{humanml3d} utilize all features including redundancy which is imprecise and unfair.

\begin{table}[t]
\renewcommand{\arraystretch}{1}
\caption{\textbf{The result with existing evaluator on HumanML3D dataset}. We alter data by adding noise or replacing it with noise in essential and redundant dimensions. The result shows the evaluator heavily emphasizes redundant dimensions during evaluation.}
\vspace{-2em}
\label{tab:diagnosis3} 
\begin{center}
\resizebox{0.8\linewidth}{!}{\begin{tabular}{c|c|c|c|c|c}
\toprule
\multirow{2}{*}{{Dimension}} & \multirow{2}{*}{{Method}} & \multicolumn{1}{c|}{FID $\downarrow$} & \multicolumn{3}{c}{R-Precision $\uparrow$} \\
 \cmidrule(r){3-6}
 & & Gen  &Top 1 &Top 2 & Top 3 \\
\midrule
     Essential & Add Noise & $2.021$ & $0.442$ & $0.634$  & $0.740$ \\
    \midrule
     Redundant & Add Noise & $21.032$ & $0.310$ & $0.471$  & $0.575$ \\
     \midrule
     Essential & Replace W$\slash$ Noise & $15.164$ & $0.264$ & $0.425$  & $0.538$ \\
    \midrule
     Redundant & Replace W$\slash$  Noise &  $38.167$ & $0.154$ & $0.257$  & $0.336$ \\
\bottomrule
\end{tabular}}
\vspace{-2.5em}
\end{center}
\end{table}
To assess this, we conducted an experiment by selectively altering redundant and non-redundant dimensions of ground truth HumanML3D data to examine their impact on the evaluator. As shown in \cref{tab:diagnosis3}, the evaluator disproportionately emphasizes redundant dimensions, potentially misclassifying accurate human motion as poor if minor imperfections exist in redundancy.

\noindent\textbf{Benefiting VQ-Based Methods.}
The VQ codebooks enforce a discrete one-to-one token-to-embedding mapping, ensuring error consistency across both essential and redundant dimensions, ultimately advantage VQ-based methods under evaluators that account for all dimensions.

Unlike traditional VAEs \cite{vae}, where the continuous latent space can lead to projected output dimensional inconsistencies, VQ-VAEs establish a one-to-one correspondence between each code and its embedding. While this approach may limit data diversity, the deterministic mapping constrains outputs to a defined set of embeddings, producing stable features across all dimensions.
Mathematically, each VQ codebook embedding $e_k$ represents a Voronoi cell:
\vspace{-2mm}
\begin{equation} 
    V_k = \{ z \in \mathbb{R}^d \mid \| z - e_k \|_2 \leq \| z - e_j \|_2\}, \forall j \neq k.
    \vspace{-2mm}
\end{equation}
Generation error then corresponds to errors of cell centroids, yielding a more uniform error rate across dimensions and aligning well with evaluators that assess all dimensions.

\noindent\textbf{Limiting Diffusion-Based Methods.}  
The continuous predictions nature of diffusion-based methods causes error inconsistency in each dimension, hindered in evaluation.

\begin{table}[t]
\renewcommand{\arraystretch}{1}
\caption{\textbf{The evaluation results using evaluators trained on all \vs essential dimensions on HumanML3D.} VQ-based models significantly outperform diffusion-based models under all-dimension evaluation, but gap closes under essential evaluation.}
\vspace{-2em}
\label{tab:diagnosis4} 
\begin{center}
\resizebox{0.8\linewidth}{!}{\begin{tabular}{c|c|c|c|c|c}
\toprule
\multirow{2}{*}{{Method}} & \multirow{2}{*}{{Evaluator With}} & \multicolumn{1}{c|}{FID $\downarrow$} & \multicolumn{3}{c}{R-Precision $\uparrow$} \\
 \cmidrule(r){3-6}
 & Redundancy & Gen  &Top 1 &Top 2 & Top 3 \\
\midrule
    T2M-GPT~\cite{t2m-gpt} & \ding{51} & $0.115$ & $0.497$ & $0.685$ & $0.779$ \\
    \midrule
    Momask~\cite{momask}  & \ding{51}  & $0.093$ & $0.508$ & $0.701$ & $0.796$ \\
    \midrule
    MDM-50Step~\cite{mdm} & \ding{51} & $0.481$  & $0.459$ & $0.651$ & $0.753$   \\
    \midrule
    T2M-GPT~\cite{t2m-gpt} & \ding{55} & $0.335$ &  $0.470$ & $0.659$  & $0.758$ \\
    \midrule
    Momask~\cite{momask} & \ding{55} & $0.116$ & $0.490$ & $0.687$  & $0.786$\\
    \midrule
    MDM-50Step~\cite{mdm}  & \ding{55} &$ 0.518$ & $0.440$ &$ 0.636$ & $ 0.741$ \\
\bottomrule
\end{tabular}}
\vspace{-2.5em}
\end{center}
\end{table}
In each timestep, diffusion-based models predict a continuous vector to recover $\mathbf{x}_0$ without dimension alignment, introducing variability across individual dimensions. As iteration continues, these dimensional fluctuations accumulate. We can express the total error with diffusion model $f$:
\vspace{-2mm}
\begin{equation}
    \text{Total Error} = \sum_{t=0}^T \sum_{d=1}^D \left\| f(\mathbf{x}_t^{(d)}, \epsilon_t^{(d)}) - \mathbf{x}_{\text{true}}^{(d)} \right\|_2^2
    \vspace{-2mm}
\end{equation}
we see each dimension contributes differently to the overall error depending on its error rate and variability across dimensions results in inconsistent error rates. Since the evaluator considers all dimensions, diffusion-based models are penalized because of error rate inconsistency, especially in the redundant dimensions, leading to unfair evaluations.

In \cref{tab:diagnosis4}, we compare model performance when evaluated on all versus essential motion dimensions. VQ-based methods (\eg, T2M-GPT) tend to outperform diffusion-based ones (\eg, MDM) under all-dimension evaluation, benefiting from more consistent error across dimensions. However, when focusing only on animation-relevant essential dimensions, MDM performs comparably to T2M-GPT, aligning better with visual perception. This suggests that diffusion-based methods, despite excelling in essential dimensions, may be penalized by fluctuations in redundancy and underscores a more robust evaluator is crucial.

\begin{figure*}[tb]
\centering
\includegraphics[width=0.9\linewidth]{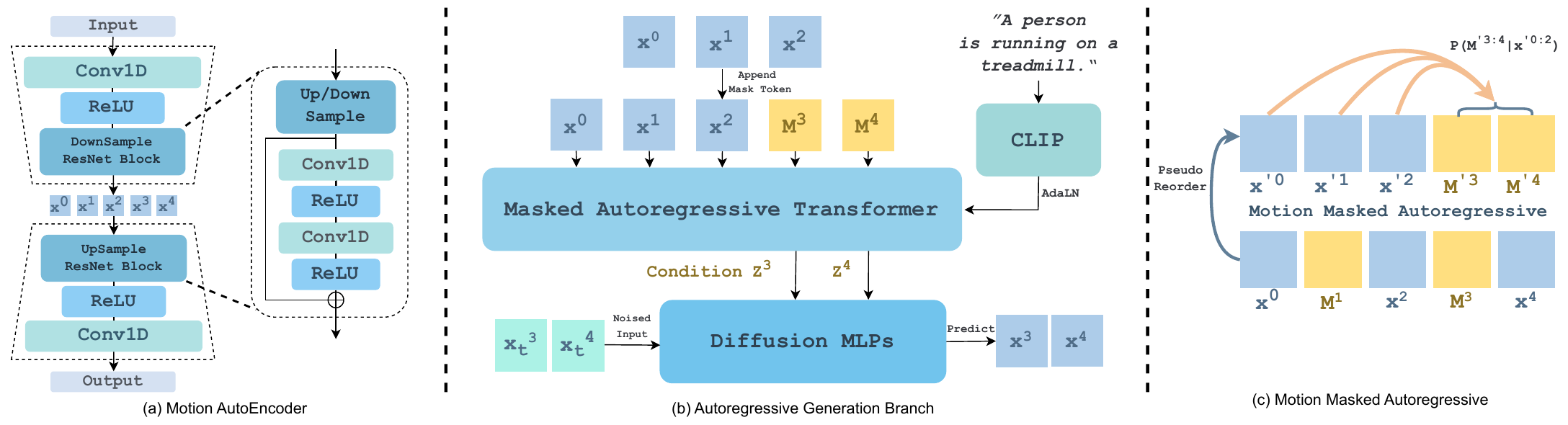}
\vspace{-1em}
\caption{\textbf{Method Overview.} \textbf{(a)} The reformed motion sequence is projected into a compact fine-grained latent space through a Motion AutoEncoder. \textbf{(b)} The motion latents $\mathbf{x}^{0:3}$ are processed through a Masked Autoregressive Transformer, where they are either randomly masked (in training) or appended (in inference) with a learnable mask vector (yellow-colored latents). The transformer provides a condition z for the masked positions to the Diffusion MLPs to produce clean latent $\mathbf{x}^{3:4}$ from the noised input. \textbf{(c)} A visual illustration of motion masked autoregressive where masked latents (yellow-colored) can be reordered into a pseudo-position allowing $p(\text{m}'^{3:4}|\mathbf{x}'^{0:2})$ prediction.}
\vspace{-0.5em}
\label{fig:architecture}
\vspace{-0.5em}
\end{figure*}
\section{Revisiting Motion Diffusion}
\label{sec:method}
Guided by the insights from \cref{sec:diagnosis} and inspirations from VQ-based methods, we revisit diffusion-based human motion generation in this section and present a new method.
It does not only overcome the limitations we found in \cref{sec:diagnosis} but also leverages the strengths of autoregressive generation, leading to a new state-of-the-art model.

\subsection{Reforming Motion Representations}
To systematically address the limitations of motion representations in \cref{sec:diagnosis1}, we only use the essential feature groups (\ie, the first $\#\text{joints}\times3+1$ dimensions).  After excluding the redundant dimensions, we avoid mixing representations from various distributions, such as 6D rotational and categorical. The retained features are all 3D continuous representations, ensuring a uniform distribution that aligns better with the diffusion-based generation framework.

To further optimize the motion representations, we then project those essential features into a compact and fine-grained latent space using a motion AutoEncoder (AE). Compared with the motion Variational AutoEncoder (VAE)~\cite{mld}, 
the deterministic AE projection avoids the variation in the motion latents (\ie $\boldsymbol{\epsilon}$ incorporation), providing more stable representations that are better suited for diffusion modeling and motion reconstruction. (In Appendix, we show training baseline methods using only essential dimensions can already lead to significant improvements, and processing into latent space may further enhance the results.)

The AE architecture is shown in the left-most part of \cref{fig:architecture}, where the motion sequence with essential representations $\mathbf{X}^{0:N}$ is projected into a latent space using a 1D ResNet~\cite{resnet}-based encoder $E$. This latent embedding $\mathbf{x}^{0:n}$ then passes through a 1D ResNet decoder $D$, which uses nearest-neighbor upsampling to reconstruct the motion feature $\mathbf{X}'^{0:N}$. Formally, the training loss of AE is defined as:
\vspace{-1mm}
\begin{equation}
    \mathcal{L}_{\text{ae}} = \|\mathbf{X}^{0:N} - \mathbf{X}^{'0:N}\|_1.
\vspace{-1mm}
\end{equation}
Following previous works~\cite{momask, t2m-gpt, mld}, the encoder $E$ downsamples $\mathbf{X}$ from $N$ length to $\mathbf{x}$ of $n$ length. The decoder $D$ upsamples it back to $N$ length. With this integration, our method can also use motion latent $\mathbf{x}$ in the diffusion process, which offers acceleration for both training and sampling.
In addition, since the downsampling introduces temporal awareness~\cite{humanml3d}, it can improve the temporal coherence of autoregressive diffusion in \cref{sec:method2}.
\newcommand{\vx}{\mathbf{x}}

More importantly, since our reformed motion representations can effectively address the issues highlighted in \cref{sec:diagnosis1}, it free diffusion models from the constraint of predicting only $\mathbf{x}_0$ to $\boldsymbol{\epsilon}$ and more advanced diffusion predictions are now feasible, \eg score and velocity. Following SiT~\cite{sit}, we define a linear interpolation function as:
\vspace{-1.5mm}
\begin{equation}
    \mathbf{x}_t = \alpha_t \mathbf{x}_0 + \sigma_t \boldsymbol{\epsilon} = (1-t) \mathbf{x}_0 + t \boldsymbol{\epsilon},
    \vspace{-1.5mm}
\end{equation}
where $t$ is continuous timestep and define velocity as:
\vspace{-1.5mm}
\begin{equation}
    \mathbf{v}(\mathbf{x},t) = \dot\alpha_t \, \mathrm{E} \left[ \mathbf{x}_0 \mid \mathbf{x}_t = \mathbf{x} \right] + \dot{\sigma}_t \, \mathrm{E} \left[ \boldsymbol{\epsilon} \mid \mathbf{x}_t = \mathbf{x} \right].
    \vspace{-1.5mm}
\end{equation}
The score is another form of velocity:
\vspace{-1.5mm}
\begin{equation}
    \mathbf{s}(\mathbf{x}, t) = \sigma_t^{-1} \frac{\alpha_t \mathbf{v}(\mathbf{x},t) - \dot \alpha_t \mathbf{x}}{\dot \alpha_t \sigma_t - \alpha_t \dot \sigma_t},
    \vspace{-1.5mm}
\end{equation}
where $\alpha_t$, $\sigma_t$ are continous time coefficients,  $\dot\alpha_t = \frac{\mathbf{d}\alpha_t}{\mathbf{d}t}$,  $\dot\sigma_t = \frac{\mathbf{d}\sigma_t}{\mathbf{d}t}$.
We can also deduce score $\mathbf{s}(\mathbf{x}, t)$ from $\mathbf{v}(\mathbf{x},t)$.

\subsection{Diffusion: An Autoregressive Approach}
\label{sec:method2}
Autoregressive methods~\cite{momask,bamm,mmm, t2m-gpt} demonstrate significant advantages in motion generation.
Instead of generating the entire sequence $\mathbf{x}^{1:n}$ with a condition \textit{c} by modeling $p(\mathbf{x}^{1:n} | c)$, autoregressive generation can simplify training and generation into a chain probability:
\vspace{-1.75mm}
\begin{equation}
p(\mathbf{x}^{1:n}|c) = p(\mathbf{x}^1 | c)\prod_{i=1}^{n} p(\mathbf{x}^i | \mathbf{x}^{<i}), \vspace{-1.75mm}
\end{equation}
where each conditional probability $p(\mathbf{x}^i | \mathbf{x}^{<i})$ represents the likelihood of generating motion $\mathbf{x}_i$ given previous ones $\mathbf{x}^{<i}$, which makes training objective significantly easier.

Naively training diffusion models to perform autoregression using MSE loss fails as it's simplified to be a regression problem rather than explicitly capturing chained probabilistic distributions of $p(\mathbf{x}^{1:n}|c)$ in autoregressive generation.

Recent advances~\cite{mar,givt} in image generation demonstrate the potential of autoregressive continuous image modeling.
They leverage logits from an autoregressive model as conditioning parameters into a continuous sampling network to better model underlying probability.
Inspired by this, we revisit human motion diffusion models from an autoregressive perspective.

\subsubsection{Masked Autoregressive Motion Generation}
We follow the masked autoregressive approach proposed by MAR~\cite{mar}. 
In each autoregressive iteration, we define unmasked motion latents as $um = \mathbf{x}^{i_1:i_k}$ and masked motion latents as $m = \mathbf{x}^{j_1:j_{n-k}}$, where $k$ is the number of autoregressive steps. The unmasked latents can be refined in a new pseudo-order (a flexible reordering process that can be sequential, random, or any custom ordering) $um = \mathbf{x}'^{1:k}$ to serve as previously generated blocks. 
The masked motion latents $m = \mathbf{x}'^{k+1:n}$ represent the motion latents that need to be generated based on $um$ with condition $c$. Formally, this process is 
\vspace{-1.5mm}
\begin{equation}
   p(\mathbf{x}'^{1:n}| c) = p(\text{m}| c) \prod_{j=1}^{k} p(\text{m}|\text{um})
    \vspace{-1.5mm}
\end{equation}
We visually illustrate motion masked autoregressive in the right-most part of \cref{fig:architecture}. Note that because of masked autoregression, our approach is capable of predicting multiple latents, not limited to two shown in \cref{fig:architecture}.

\subsubsection{Autoregressive Generation Branch}
Our proposed autoregressive diffusion generation architecture is shown in the middle of \cref{fig:architecture}.
It consists of two major parts: a Masked Autoregressive Transformer and per-latent Diffusion Multi-Layer Perceptions (MLPs).

\noindent \textbf{Masked Autoregressive Transformer} is designed to process and understand time-variant motion data, and provide rich contextual condition $z$ for the diffusion branch. Given unmasked motion latent sequence $um$ as previous generated motion latents, the masked autoregressive transformer $g$ will produce conditions $z$ to the diffusion branch to generate latents at position of masked motion latents $m$ by:
\vspace{-1.5mm}
\begin{equation}
    z = g(\text{um}).
    \vspace{-1.5mm}
\end{equation}
Given the relative simplicity of motion data compared to image data, we only use one single AdaLN~\cite{adaln} transformer~\cite{transformer} layer.
To balance the generation performance and speed, we use bidirectional attention, same as many previous methods~\cite{mmm, bamm, momask}.

\noindent \textbf{Diffusion MLPs} adopts MLPs as its primary structure because with autoregressive modeling, the motion data input into the diffusion branch is independent single $D$-dimensional motion latent.  
The single $D$-dimensional structure input data aligns well with the MLP's simplicity and strength in channel-wise manipulation. 
Unlike MAR~\cite{mar}, which scales the understanding models, we scale generative diffusion MLPs to various model sizes. 
Using the condition $z$ from Masked Autoregressive Transformer, diffusion branch produce each motion latents $\mathbf{x}'^{i}$ in masked token $m$'s position at each timestep $t$ by:
\vspace{-1.5mm}
\begin{equation}
    \mathbf{x}'^{i}_{t-1} \sim p(\mathbf{x}'^{i}_{t-1}|\mathbf{x}'^{i}_{t}, t, z^{i})
    \vspace{-1.5mm}
\end{equation}
During training, we randomly mask a subset of $k$ motion latents with a learnable continuous mask vector following the cosine masking schedule from MoMask~\cite{momask}. The autoregressive model learns to provide accurate signals based on the unmasked latents given time step and text condition.
The diffusion MLPs utilize this signal to predict $\boldsymbol{\epsilon}$ or $\mathbf{v}(\mathbf{x},t)$ on masked latents.
Training objectives for the entire generation branch are denoted as:
\vspace{-2mm}
\begin{equation}
    \mathcal{L}_{GB} = \mathbb{E}_{\boldsymbol{\epsilon},t}\left[{ \left\| \boldsymbol{\epsilon} - \boldsymbol{\epsilon}_\theta(\mathbf{x'}_t^{i}|t, g(\text{um}) \right\|^2}\right]
    \vspace{-2mm}
\end{equation}
For velocity prediction, $\mathcal{L}_{GB}=$
\vspace{-2mm}
\begin{equation}
    \int_0^T \mathbb{E}_{\mathbf{v},t}\left[\| \mathbf{v}_\theta(\mathbf{x'}_t^{i}|t, g(\text{um}) - \dot\alpha_t \mathbf{x'}^{i}_0 - \dot\sigma_t \boldsymbol{\epsilon}\|^2\right]\mathrm{d}t
     \vspace{-2mm}
\end{equation}
\begin{table*}[t]
    \centering
    \renewcommand{\arraystretch}{1.0}
    \vspace{-0.5em}
    \resizebox{0.9\linewidth}{!}{
    \begin{tabular}{l|l| l| c c c| c |c| c| c}
    \toprule
      & \multirow{2}{*}{Methods} & \multirow{2}{*}{Framework}  & \multicolumn{3}{c|}{R-Precision$\uparrow$} & \multirow{2}{*}{FID$\downarrow$} & \multirow{2}{*}{Matching$\downarrow$} & \multirow{2}{*}{MModality$\uparrow$} & \multirow{2}{*}{CLIP-score$\uparrow$}\\
    \cline{4-6}
       ~& ~ & ~& Top 1 & Top 2 & Top 3 & & & & \\
    \midrule
    \multirow{9}{*}{\rotatebox{90}{\textbf{HumanML3D}}} &
    T2M-GPT~\cite{t2m-gpt} & \multirow{3}{*}{VQ} & $0.470^{\pm.003}$ &$0.659^{\pm.002}$  & $0.758^{\pm.002}$ &$0.335^{\pm.003}$ & $3.505^{\pm.017}$& $2.018^{\pm.053}$ & $0.607^{\pm.005}$ \\
    & MMM~\cite{mmm} & &$0.487^{\pm.003}$&$0.683^{\pm.002}$&$0.782^{\pm.001}$&$0.132^{\pm.004}$&$3.359^{\pm.009}$& $1.241^{\pm.073}$ & $0.635^{\pm.003}$ \\ 
    & MoMask~\cite{momask} & &$0.490^{\pm.004}$ & $0.687^{\pm.003}$  &$0.786^{\pm.003}$ & $0.116^{\pm.006}$ &$3.353^{\pm.010}$& $1.263^{\pm.079}$ & $\underline{0.637^{\pm.003}}$\\
    \cline{2-10}
    & MDM-50Step~\cite{mdm} & \multirow{4}{*}{Diffusion} &$0.440^{\pm.007}$ & $0.636^{\pm.006}$& $0.742^{\pm.004}$& $0.518^{\pm.032}$ & $3.640^{\pm.028}$& $\mathbf{3.604^{\pm.031}}$ & $0.578^{\pm.003}$\\
    & MotionDiffuse~\cite{motiondiffuse} & &$0.450^{\pm.006}$ &$0.641^{\pm.005}$&$0.753^{\pm.005}$&$0.778^{\pm.005}$&$3.490^{\pm.023}$&$3.179^{\pm.046}$ & $0.606^{\pm.004}$\\
    & MLD~\cite{mld} & &$0.461^{\pm.004}$ &$0.651^{\pm.004}$&$0.750^{\pm.003}$&$0.431^{\pm.014}$&$3.445^{\pm.019}$&$\underline{3.506^{\pm.031}}$ & $0.610^{\pm.003}$\\
    & ReMoDiffuse~\cite{remodiffuse} & &$0.468^{\pm.003}$& $0.653^{\pm.003}$& $0.754^{\pm.005}$ & $0.883^{\pm.021}$&$3.414^{\pm.020}$& $2.703^{\pm.154}$&$0.621^{\pm.003}$\\
    \cline{2-10}
    & \textbf{Ours-DDPM} & Autoregressive & $\underline{0.492^{\pm.006}}$ & $\underline{0.690^{\pm.005}}$& $\underline{0.790^{\pm.005}}$& $\underline{0.116^{\pm.004}}$ & $\underline{3.349^{\pm.010}}$& $2.470^{\pm.053}$ & $0.637^{\pm.005}$ \\
    & \textbf{Ours-SiT} & Diffusion& $\mathbf{0.500^{\pm.004}}$ & $\mathbf{0.695^{\pm.003}}$& $\mathbf{0.795^{\pm.003}}$& $\mathbf{0.114^{\pm.007}}$ & $\mathbf{3.270^{\pm.009}}$& $2.231^{\pm.071}$ & $\mathbf{0.642^{\pm.002}}$ \\
    
    \midrule
    \multirow{9}{*}{\rotatebox{90}{\textbf{KIT}}} &
     T2M-GPT~\cite{t2m-gpt} &\multirow{3}{*}{VQ} &$0.359^{\pm.007}$&$0.553^{\pm.007}$&$0.690^{\pm.013}$&$0.593^{\pm.053}$& $3.765^{\pm.046}$& $1.798^{\pm.157}$& $0.651^{\pm.005}$\\
     & MMM~\cite{mmm} &&$0.363^{\pm.005}$&$0.569^{\pm.006}$&$0.724^{\pm.006}$&$0.478^{\pm.034}$& $3.629^{\pm.028}$& $1.455^{\pm.106}$& $0.660^{\pm.003}$\\ 
     &MoMask~\cite{momask} && $0.369^{\pm.005}$ & $0.588^{\pm.005}$& $0.731^{\pm.005}$& $0.411^{\pm.026}$ & $3.577^{\pm.021}$& $1.309^{\pm.058}$ & $0.669^{\pm.002}$\\
    \cline{2-10}
     &MDM~\cite{mdm} & \multirow{4}{*}{Diffusion} &$0.333^{\pm.012}$ & $0.561^{\pm.009}$& $0.689^{\pm.009}$& $0.585^{\pm.043}$ & $4.002^{\pm.033}$ & $1.681^{\pm.107}$ &$0.605^{\pm.007}$\\
     &MotionDiffuse~\cite{motiondiffuse} &&$0.344^{\pm.009}$& $0.536^{\pm.007}$&$0.658^{\pm.007}$&$3.845^{\pm.087}$& $4.167^{\pm.054}$&$1.774^{\pm.217}$&$0.626^{\pm.006}$\\
     &MLD~\cite{mld} &&$0.351^{\pm.007}$& $0.536^{\pm.007}$&$0.658^{\pm.007}$&$0.492^{\pm.047}$& $3.746^{\pm.044}$&$\underline{1.803^{\pm.164}}$&$0.646^{\pm.006}$\\
     &ReMoDiffuse~\cite{remodiffuse} &&$0.356^{\pm.004}$&$0.572^{\pm.007}$& $0.706^{\pm.009}$&$1.725^{\pm.053}$&$3.735^{\pm.036}$&$\mathbf{1.928^{\pm.127}}$& $0.665^{\pm.005}$\\
    \cline{2-10}
     &\textbf{Ours-DDPM} &Autoregressive& $\underline{0.375^{\pm.006}}$ & $\underline{0.597^{\pm.008}}$& $\underline{0.739^{\pm.006}}$& $\underline{0.340^{\pm.020}}$ & $\underline{3.489^{\pm.018}}$& $1.479^{\pm.078}$ & $\underline{0.681^{\pm.003}}$\\
     &\textbf{Ours-SiT} &Diffusion& $\mathbf{0.387^{\pm.006}}$ & $\mathbf{0.610^{\pm.006}}$& $\mathbf{0.749^{\pm.006}}$& $\mathbf{0.242^{\pm.014}}$ & $\mathbf{3.374^{\pm.019}}$& $1.312^{\pm.053}$&$\mathbf{0.692^{\pm.002}}$\\
    \bottomrule
    \end{tabular}}
    \caption{\textbf{Quantitative evaluation on HumanML3D and KIT-ML datasets.} We repeat the evaluation 20 times and report the average with 95\% confidence interval. For our methods, we report both method results trained to predict noise (DDPM\cite{ddpm}) and velocity (SiT\cite{sit}). We use \textbf{Bold} face to indicate the best result and \underline{underscore} to present the second best.
    }
    \label{tab:result}
    \vspace{-1em}
\end{table*}
During sampling, given previous latents $um$, we simply add mask vectors to the sequence, allowing the autoregressive model to generate appropriate signals for masked positions. 
The sampling process can be denoted as:
\vspace{-2mm}
\begin{equation}
    \mathbf{x}_{t-1}^i = \frac{1}{\sqrt{\alpha_t}} \left( \mathbf{x}_t^i - \frac{\sqrt{1 - \alpha_t}}{\sqrt{1 - \bar{\alpha}_t}} \, \boldsymbol{\epsilon}_\theta(\mathbf{x}_t^i \mid t, z^i) \right) + \sigma_t \boldsymbol{\epsilon}_t
    \vspace{-2mm}
\end{equation}
where $\boldsymbol{\epsilon}_t\sim \mathcal{N}(\mathbf{0}, \mathbf{I})$ for noise prediction. The velocity prediction uses ODE sampling with step size $\Delta t$:
\vspace{-2mm}
\begin{equation}
    \mathbf{x}_{t-1}^i = \mathbf{x}_t^i + \Delta t \cdot \mathbf{v}_\theta(\mathbf{x}_t^i\mid t, z^i)
    \vspace{-1mm}
\end{equation}

\subsection{Evaluation: More Robust Evaluators.}
\label{sec:method3}
To address biases in the current evaluation approach shown in \cref{sec:diagnosis2} due to the unnecessary focus on redundant motion representations, we propose a new evaluation framework that focuses exclusively on essential features (the only ones meaningful), enabling a fairer generation evaluation.

We first construct an evaluator that retains the architecture of the widely used evaluator~\cite{humanml3d}, which consists of a convolutional movement encoder, a GRU~\cite{gru}-based motion encoder, and a GRU-based text encoder using GloVe~\cite{glove} embeddings.
This evaluator is trained solely on essential dimensions that meaningfully contribute to final motion generation with no other alterations.

To incorporate recent advancements, we also design a CLIP~\cite{clip, intergen, motionclip}-based evaluator trained with per-batch contrastive learning. Specifically, motion captions are tokenized, embedded, and processed through a transformer encoder branch, while motion data are projected and processed through another transformer encoder branch. The end-of-sentence token from the text embeddings and the CLS token from the motion embeddings are extracted to represent each modality. The model learns to align these representations by maximizing the per-batch cosine similarity between the normalized features of two modalities scaled by a learnable logit scale. This CLIP-based evaluator is also trained using only essential dimensions.

By training exclusively on essential dimensions, we ensure the evaluators capture only meaningful features from the final generated motion. The adoption of dual evaluators also provides a more robust and comprehensive framework to accurately compare different-based generation methods. More importantly, all baselines can generate outputs containing essential dimensions with no additional operations.
\vspace{-1.5em}
\section{Experiment}
\label{sec:experiment}
\begin{figure*}[tb]
\centering
\includegraphics[height=3.25cm, width=0.85\linewidth]{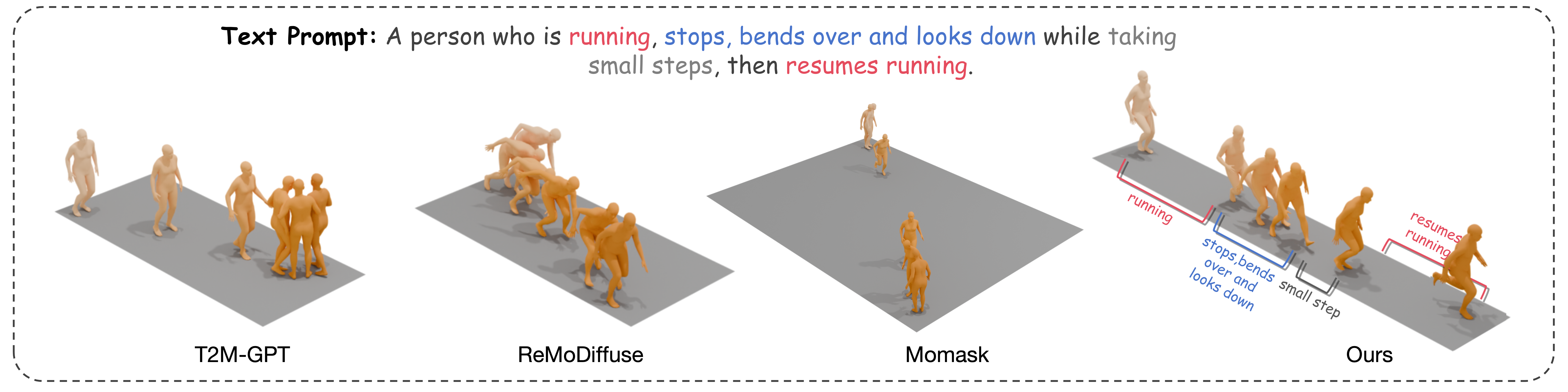}
\vspace{-1em}
\caption{\textbf{Visualization Comparison} between our method and baseline state-of-the-art methods. Our method generates motion that is more realistic and more accurately follows the fine details of the textual condition.}
\label{fig:qualitative}
\vspace{-1em}
\end{figure*}

\subsection{Datasets and Evaluation Protocols.}
\vspace{-0.25em}
\noindent\textbf{Datasets.}
\vspace{-0.25em}
To accurately and fairly evaluate our method in comparison with
baselines, we adopts two representative motion-language benchmarks: HumanML3D~\cite{humanml3d} and KIT-ML~\cite{kit}. 
The KIT-ML dataset comprises 3,911 motions sourced from the KIT and CMU~\cite{cmu} motion data, each accompanied with one to four textual annotations (6,278 total annotations). The KIT-ML motion sequences are standardized to 12.5 FPS.
The HumanML3D dataset contains 14,616 motions sourced from the AMASS~\cite{amass} and HumanAct12~\cite{humanact12} datasets, each described by three textual scripts (44,970 total annotations). The HumanML3D motion sequences are adjusted to 20 FPS with a maximum duration of 10 seconds.
We augment data by mirroring and splitting both datasets into train, test, and validation sets with a ratio of 0.8:0.15:0.05. We follow the pose representation from T2M~\cite{humanact12}, however, we incorporate only essential dimensions for methods' evaluation and training in our method.

\noindent\textbf{Evaluation Metrics.}
Following \Cref{sec:method3}, we employ two evaluators trained with only essential dimensions: one architecturally identical to the one proposed in T2M~\cite{humanml3d} and a CLIP-based evaluator.
Using the T2M evaluator, we adopt evaluation metrics from T2M, including (1) R-Precision (Top-1, Top-2, and Top-3 accuracies) and Matching, which measures the semantic alignment between generated motion embeddings and their corresponding captions' glove embedding; (2) Fréchet Inception Distance (FID), which assesses the statistical similarity between ground truth and generated motion distributions; and (3) MultiModality, which measures the diversity of generated motion embeddings per same text prompt.
Using the CLIP-based evaluator, we include the CLIP-score~\cite{clipscore}, which measures the compatibility of motion-caption pairs by calculating the cosine similarity between generated motion and its caption.

\subsection{Results and Analysis}
\vspace{-0.25em}
Following previous works~\cite{humanml3d, mdm}, we conduct each experiment 20 times on both datasets and report the mean result along with a 95\% confidence interval. To ensure a fair comparison on the new evaluators, we train all baseline models from scratch following their original methods on the same obtained dataset (full dimension). We present the quantitative results of our method alongside baseline state-of-the-art human motion generation methods in \cref{tab:result}, and the qualitative comparison results in \cref{fig:qualitative}.
In addition we also present model scalability results in App. \ref{app:additional_quantitative}

As observed, our method achieves superior performance across multiple metrics, including FID, R-Precision, Matching score, and CLIP-score, consistently outperforming baseline methods with non-marginal improvements on both KIT-ML and HumanML3D datasets.
Compared to diffusion-based baseline methods, our approach showcases a significantly stronger ability to generate stable motion that follows closely to text instructions and aligns with ground truth. Notably, while the SOTA diffusion-based baseline method ReMoDiffuse~\cite{remodiffuse} relies on additional data retrieval from a large database to achieve high-quality motion generation, our method surpasses ReMoDiffuse's performance results without requiring auxiliary formulations.
In comparison to VQ-based baseline methods, our approach maintains better motion quality and also delivers greater diversity, a quality where VQ methods underperform.
\vspace{-0.25em}
\subsection{Ablation Study}
\vspace{-0.25em}
\begin{table}[t]
\renewcommand{\arraystretch}{1.0}
\caption{\textbf{Ablation study results} comparing our method to variations without reform data representation and distribution and without autoregression. The study is conducted on the HumanML3D dataset.}
\vspace{-1.5em}
\label{tab:ablation} 
\begin{center}
\resizebox{0.95\linewidth}{!}{\begin{tabular}{cc|c|c|c|c}
\toprule
\multicolumn{2}{c|}{\multirow{2}{*}{{Method}}} & \multirow{2}{*}{FID $\downarrow$} & \multicolumn{3}{c}{R-Precision $\uparrow$}\\
 \cmidrule(r){4-6}
 & &  &Top 1 &Top 2 & Top 3 \\
\midrule
    \multicolumn{2}{c|}{Full Components} & $0.116$& $0.492$ & $0.690$ & $0.790$ \\
    \midrule
     \multicolumn{2}{c|}{w/o Motion Representation Reformation} & $2.196$ & $0.387$ & $0.595$ & $0.703$ \\
    \midrule
      \multicolumn{2}{c|}{w/o Autoregression} & $0.551$ & $0.435$& $0.621$ & $0.732$\\
\bottomrule
\end{tabular}}
\vspace{-2.5em}
\end{center}
\end{table}
In the ablation study, we further study the impact of reforming data representation and distribution, and autoregressive modeling in our method. We present ablation results in \cref{tab:ablation} and an optimization routine in Appendix. The results shows that both proposed components contribute greatly.

\noindent\textbf{Data Representation and Distribution}
Without reform motion data representation and distribution, FID increased by 2.080, and R-precisions dropped by 8.7 percent for Top 3.
Therefore, reforming data and representation is crucial.

\noindent\textbf{Autoregressive Modeling}
Without autoregressive modeling, FID increased by 0.435, and R-precisions dropped by 5.8 percent for Top 3. Therefore autoregressive modeling also contributes greatly to our proposed method.

% \vspace{-0.5em}
\section{Related Work}
% \vspace{-0.25em}
\label{sec:related_work}
In this section, we provide a brief overview of related works due to space constraints. We also provide a more detailed version of related works in the Appendix.

\noindent\textbf{VQ-Based Human Motion Generation}
TM2T~\cite{tm2t} first introduces Vector Quantization (VQ) to text-to-human motion generation, enabling discrete motion token modeling. T2M-GPT~\cite{t2m-gpt} extended this by leveraging a GPT\cite{gpt} to motion autoregressive generation. Subsequent methods have sought to integrate a larger model\cite{jiangmotiongpt, zhangmotiongpt} (\eg large language models), or manipulate attention mechanisms ~\cite{attt2m}.
Most recently, MMM~\cite{mmm} and MoMask~\cite{momask} revisit generation methodology by employing bidirectional masked generation techniques inspired by MaskGIT~\cite{maskgit}. In this paper, we examine the strengths of these methods and improve a diffusion model with these insights.

\noindent\textbf{Diffusion-Based Human Motion Generation.}
Inspired by the success of denoising diffusion models in image generation domain~\cite{ddpm, ddim}, several pioneering works~\cite{mdm, flame, motiondiffuse} have adapted denoising diffusion processes to human motion generation. Building on these works, MLD~\cite{mld}further optimized the denoising process in latent space to improve training and sampling efficiency.  Recent methods have diversified their focus, exploring retrieval-augmention~\cite{remodiffuse}, controllable generation~\cite{motionlcm}, as well as investigating advanced architectures~\cite{motionmamba} such as Mamba~\cite{mamba}. In this paper, we thoroughly investigate the limitations of diffusion-based methods and propose to address them.

\noindent\textbf{Autoregressive Generation with Continous Data.}
GIVT first introduced the idea of leveraging outputs from an autoregressive model as parameters for a Gaussian Mixture Model, enabling probabilistic autoregressive modeling and generation.
MAR then utilized logits from a masked autoregressive model as input to a small diffusion branch, producing more fine-grained generation. Inspired by these approaches, in this paper, we propose a novel framework that integrates diffusion-based motion generation with autoregression to achieve enhanced generative performance.

% \vspace{-0.75em}
\section{Conclusion}
\label{sec:conclusion}
% \vspace{-0.25em}
In conclusion, we introduce a novel diffusion-based generative framework for text-driven 3D human motion generation. Our method reforms motion data representation and distribution to better fit the diffusion model, incorporates masked autoregressive training and sampling techniques and is evaluated by more robust evaluators. Extensive experiments demonstrate our method’s superior generation performance in KIT-ML and HumanML3D datasets.

\vskip4pt \noindent{\bf Acknowledgement.~}
Yiming Xie was supported by the Apple Scholars in AI/ML PhD fellowship.

{
    \small
    \bibliographystyle{ieeenat_fullname}
    \bibliography{main}
}

% WARNING: do not forget to delete the supplementary pages from your submission 
\clearpage
\appendix
\maketitlesupplementary

\setcounter{section}{0}
\setcounter{table}{0}
\renewcommand{\thetable}{A\arabic{table}}
\renewcommand*{\theHtable}{\thetable}

\setcounter{figure}{0}
\renewcommand{\thefigure}{A\arabic{figure}}
\renewcommand*{\theHfigure}{\thefigure}

We further discuss our proposed approach with the following supplementary materials:
\begin{itemize}
\item \cref{app:proof}: Detailed Deduction
\item \cref{app:related_work}: Detailed Related Works
\item \cref{app:implementation_details}: Implementation Details
\item \cref{app:additional_quantitative}: Additional Quantitative Results
\item \cref{app:editing_inpainting}: Temporal Editing
\item \cref{app:additional_qualitative}: Additional Qualitative Results
\item \cref{app:limitation}: Limitations
\end{itemize}

\section{Detailed Deduction}
\label{app:proof}
\subsection{Detailed Deduction for \cref{equation4}}
\label{app:proof1}
In paper, we define $\delta_{\mathbf{x}_0}$ and $\delta_{\boldsymbol{\epsilon}}$ to be:
\begin{equation}
    \delta_{\boldsymbol{\epsilon}} = \| \boldsymbol{\epsilon}_\theta(\mathbf{x}_t, t) - \boldsymbol{\epsilon} \|_2^2
\end{equation}
and
\begin{equation}
    \delta_{\mathbf{x}_0} = \left\| \mathbf{x}'_0 - \mathbf{x}_0 \right\|_2^2
\end{equation}
Since in diffusion-based methods, in each step, diffusion-based methods reconstruct the original motion by:
\begin{equation}
    {\mathbf{x}_0}' = \frac{1}{\sqrt{\bar{\alpha}_t}}(\mathbf{x}_t - \sqrt{1 - \bar{\alpha}_t}\boldsymbol{\epsilon}_\theta(\mathbf{x}_t, t))
\end{equation}
where $\boldsymbol{\epsilon}_\theta(\mathbf{x}_t, t)$ is the model's prediction of the noise $\boldsymbol{\epsilon}$.
Then we have:
\begin{equation}
     \delta_{\mathbf{x}_0} =  \left\| \frac{1}{\sqrt{\bar{\alpha}_t}} \left(\mathbf{x}_t - \sqrt{1 - \bar{\alpha}_t} \boldsymbol{\epsilon}_\theta(\mathbf{x}_t, t)\right) - \mathbf{x}_0 \right\|_2^2 \notag
\end{equation}
If we substitute $\mathbf{x}_0$ from \cref{equation1}:
\begin{align}
    \delta_{\mathbf{x}_0} &= \left\| \frac{1}{\sqrt{\bar{\alpha}_t}} \left( \sqrt{\bar{\alpha}_t} \mathbf{x}_0 + \sqrt{1 - \bar{\alpha}_t} \boldsymbol{\epsilon} \right) \right.\\
    &\quad\quad \left. - \frac{1}{\sqrt{\bar{\alpha}_t}} \left( \sqrt{1 - \bar{\alpha}_t} \boldsymbol{\epsilon}_\theta(\mathbf{x}_t, t) \right) - \mathbf{x}_0 \right\|_2^2 \nonumber \\ \notag
    &= \left\| \mathbf{x}_0 + \frac{\sqrt{1 - \bar{\alpha}_t}}{\sqrt{\bar{\alpha}_t}} \left( \boldsymbol{\epsilon} - \boldsymbol{\epsilon}_\theta(\mathbf{x}_t, t) \right) - \mathbf{x}_0\right\|_2^2 \\ \notag
    &= \left\| \frac{\sqrt{1 - \bar{\alpha}_t}}{\sqrt{\bar{\alpha}_t}} \left( \boldsymbol{\epsilon} - \boldsymbol{\epsilon}_\theta(\mathbf{x}_t, t) \right) \right\|_2^2 \\ \notag
    &= \left\| \frac{\sqrt{1 - \bar{\alpha}_t}}{\sqrt{\bar{\alpha}_t}}\right\|_2^2 \delta_{\boldsymbol{\epsilon}}
\end{align}
an standard error relation $\delta_{\boldsymbol{\epsilon}} \to \delta_{\mathbf{x}_0}$ if $\mathbf{x}_0$ is processed correctly which should only responds to time coefficient $\bar{\alpha}$.

\section{Detailed Related Works}
\label{app:related_work}
\noindent\textbf{Human Motion Generation.}
Early text-to-motion approaches~\cite{ahuja2019language2pose, humanml3d,petrovich2022temos,petrovich2023tmr,motionclip,yan2023cross} attempt to align the latent spaces of text and motion. 
However, this strategy encounters significant challenges in generating high-fidelity motions due to the inherent difficulty of seamlessly aligning these fundamentally distinct latent spaces. 
Consequently, recent advancements in human motion generation have shifted focus toward diffusion-based and VQ-based methods, as discussed below.

\noindent\textbf{Diffusion-Based Human Motion Generation.}
Inspired by the success of denoising diffusion models in the image generation domain~\cite{ddpm, ddim}, several pioneering works~\cite{mdm, flame, motiondiffuse} have adapted denoising diffusion processes to human motion generation. 
Building on these works, MLD~\cite{mld} further optimized the denoising process in latent space to improve training and sampling efficiency.  
PhysDiff~\cite{yuan2023physdiff} added the physical constraints in the motion generation.
And a lot of following works~\cite{azadi2023make,liang2024omg,pi2024motion,kapon2024mas,zhang2024energymogen,andreou2024lead,li2024lifting,fu2024mogo,hu2023motion,wan2023diffusionphase,zhou2025emdm,lou2023diversemotion} keep exploring diffusion-based human motion generation from different perspectives.
In this paper, we thoroughly investigate the limitations of diffusion-based methods and propose a novel approach to address them.

\noindent\textbf{VQ-Based Human Motion Generation.}
TM2T~\cite{tm2t} first introduces Vector Quantization (VQ) to text-to-human motion generation, enabling discrete motion token modeling. 
A lot of the following works~\cite{t2m-gpt,yuan2024mogents,mmm,momask,maskgit,bamm,attt2m,li2024lamp} improved the VQ-based methods.
T2M-GPT~\cite{t2m-gpt} extended this by leveraging a GPT~\cite{gpt} to motion autoregressive generation. Subsequent methods have sought to integrate a larger model~\cite{jiangmotiongpt, zhangmotiongpt} (\eg large language models), or manipulate attention mechanisms ~\cite{attt2m}.
Most recently, MMM~\cite{mmm} and MoMask~\cite{momask} revisit generation methodology by employing bidirectional attention-based masked generation techniques inspired by MaskGIT~\cite{maskgit}. BAMM~\cite{bamm} introduced a dual-iteration framework that combines unidirectional generation with bidirectional refinement to enhance the coherence of generated motions. 
The concurrent work ScaMo~\cite{lu2024scamo} explored the scaling law in human motion generation by training the model with large-scale data.
In this paper, we examine the strengths of these approaches and improve a diffusion model inspired by these insights.

\noindent\textbf{Autoregressive Generation with Continous Data.}
In motion synthesis, recent works~\cite{tedi, CAMDM, A-MDM, dart, closd} have started to explore integrating autoregressive structures into diffusion-based frameworks. However, due to the challenges of performing direct causal next motion prediction with MSE loss (as done in discrete token settings), these methods typically only use previously generated motion as a prefix condition, rather than modeling the next step motion directly using previous motion as input.
In contrast, recent image generation methods have explored tighter coupling between autoregression and diffusion. GIVT introduced the idea of giving previous generation as input, using outputs from an autoregressive model as parameters for a Gaussian Mixture Model to enable probabilistic chaining of autoregressive generation. MAR further refined this by feeding logits from a masked autoregressive model into a small diffusion branch, producing more fine-grained generation.
Inspired by these approaches, we propose to integrate diffusion-based motion generation with masked autoregression, enabling a more direct autoregressive technique beyond simple prefix conditioning to achieve improved generative performance.

\noindent\textbf{Human Motion Generation and Beyond.}
Recent methods have diversified their focus, exploring retrieval-augmention~\cite{remodiffuse}, controllable generation~\cite{xieomnicontrol,motionlcm,kaufmann2020convolutional,karunratanakul2023gmd,rempeluo2023tracepace,wan2023tlcontrol,pinyoanuntapong2024controlmm}, human-scene/object interactions~\cite{peng2023hoi,huang2023diffusion,kulkarni2023nifty,xu2023interdiff,Pi_2023_ICCV,li2023object,chen2024sitcom,wang2024sims,gong2024diffusion,yi2025generating,xu2024interdreamer,ma2024contact,cong2024laserhuman,wu2024thor,jiang2024scaling,li2025controllable,liu2025revisit,diller2024cg,Zhao:ICCV:2023,Wang_2022}, human-human interaction~\cite{javed2024intermask,xu2024inter,wang2023intercontrol,ghosh2023remos,liang2024intergen,cenready}, stylized human motion generation~\cite{zhong2025smoodi,guo2024generative,li2024mulsmomultimodalstylizedmotion}, more datasets~\cite{xu2024motionbank,lin2023motionx}, long-motion generation~\cite{zhuo2024infinidreamer,petrovich2024multi}, voice-conditioned motion generation~\cite{chen2024language}, unified motion generation and understanding~\cite{li2024unimotion}, shape-aware motion generation~\cite{tripathi2025humos}, fine-grained text controlled generation~\cite{zou2025parco,huang2025controllable,yazdian2023motionscript,shi2023generating,jin2024act}, fine-tuning pretrained motion generation model as priors~\cite{karunratanakul2024optimizing,shafir2023human}, as well as investigating advanced architectures~\cite{motionmamba,wang2024text} such as Mamba~\cite{mamba}. 

\begin{table}[t]
\renewcommand{\arraystretch}{1.5}
\caption{\textbf{Reconstruction Results} of latent encoders in our method vs baseline methods on HumanML3D~\cite{humanml3d} data. The AutoEncoder in our method exhibits better reconstruction results.}
\vspace{-1em}
\label{tab:ae}
\begin{center}
\resizebox{\linewidth}{!}{%
\begin{tabular}{c|c|c|c|c|c}
\toprule
\multirow{2}{*}{Methods} & \multirow{2}{*}{FID $\downarrow$} & \multirow{2}{*}{MPJPE $\downarrow$} & \multicolumn{3}{c}{R-Precision $\uparrow$} \\
\cmidrule(r){4-6}
 & & & Top 1 & Top 2 & Top 3 \\
\midrule 

VQ-VAE~\cite{t2m-gpt}  & $0.081^{\pm.001}$ & $72.6^{\pm.001}$ &$0.483^{\pm.003}$ & $0.680^{\pm.003}$ & $0.780^{\pm.002}$ \\
\midrule
RVQ-VAE~\cite{momask} & $0.029^{\pm.001}$ & $31.5^{\pm.001}$ &$0.497^{\pm.002}$ & $0.693^{\pm.003}$ & $0.791^{\pm.002}$ \\
\midrule
VAE~\cite{mld}& $0.023^{\pm.001}$ & $13.7^{\pm.001}$& $0.499^{\pm.002}$ & $0.695^{\pm.003}$ & $0.791^{\pm.003}$ \\
\midrule
AE (Ours) & $\mathbf{0.004^{\pm.001}}$ & $\mathbf{1.0^{\pm.001}}$ & $\mathbf{0.502^{\pm.003}}$ & $\mathbf{0.696^{\pm.002}}$ & $\mathbf{0.793^{\pm.002}}$ \\
\bottomrule
\end{tabular}}
\vspace{-1.5em}
\end{center}
\end{table}
\begin{table}
  \centering
  \caption{Further Ablation Study and Optimization Routine.}
   \vspace{-0.3cm}
  \label{tab:optmize_routine}
  \resizebox{0.85\linewidth}{!}{
  \begin{tabular}{l c| c |c | c}
    \hline
    \multirow{2}{*}{Approach} & \multirow{2}{*}{FID$\downarrow$} & \multicolumn{3}{c}{R-Precision$\uparrow$}\\
    && Top-1 & Top-2 & Top-3\\
    \hline
    MDM~\cite{mdm}-50Step-$\boldsymbol{\epsilon}$ & $31.265$ &$0.054$ &$0.103$  & $0.147$ \\
    +Masked AR & $2.196$ & $0.387$ & $0.595$ & $0.703$ \\
    ++Essential Only & $0.657$ & $0.475$ & $0.668$ & $0.774$ \\
    \cline{1-5}
    \textbf{+++AE (Ours)} & $\mathbf{0.116}$ & $\mathbf{0.492}$ & $\mathbf{0.690}$& $\mathbf{0.790}$\\
    ++++$\mathbf{X}_0$-Pred & $0.135$ &$0.485$ &$0.686$  & $0.784$\\
    \textbf{++++Velocity  (Ours)} & $\mathbf{0.114}$ & $\mathbf{0.500}$ & $\mathbf{0.695}$& $\mathbf{0.795}$\\
    \hline
  \end{tabular}
  }
      \vspace{-0.25cm}
\end{table}
\begin{table}
  \centering
  \caption{Training Baseline Methods with Reformed Motion Data Representation and Distribution, Linear schedule and $\boldsymbol{\epsilon}$-prediction}
  \vspace{-0.25cm}
  \label{tab:mdm_new}
  \resizebox{0.85\linewidth}{!}{
  \begin{tabular}{l c| c |c | c}
    \hline
    \multirow{2}{*}{Approach} & \multirow{2}{*}{FID$\downarrow$} & \multicolumn{3}{c}{R-Precision$\uparrow$}\\
    && Top-1 & Top-2 & Top-3\\
    \hline
    MDM~\cite{mdm} & $1.574$ & $0.279$ & $0.336$ & $0.415$\\
    MDM~\cite{mdm}-Essential & $0.753$ & $0.436$ & $0.627$ & $0.732$\\
    MotionDiffuse~\cite{motiondiffuse}&$0.778$ &$0.450$&$0.641$&$0.753$\\
    MotionDiffuse~\cite{motiondiffuse}-Essential & $0.533$ & $0.459$ & $0.650$ & $0.757$\\
    MDM~\cite{mdm}-Latent& $0.327$ & $0.475$ & $0.663$ & $0.768$\\
    \hline
  \end{tabular}
  }
    \vspace{-0.25cm}
\end{table}
\begin{table}
  \centering
  \caption{Original Evaluator Results on HumanML3D.}
  \vspace{-0.4cm}
  \label{tab:original_evaluator}
  \resizebox{0.75\linewidth}{!}{\begin{tabular}{l c| c |c | c}
    \hline
    \multirow{2}{*}{Approach} & \multirow{2}{*}{FID$\downarrow$} & \multicolumn{3}{c}{R-Precision$\uparrow$}\\
    && Top-1 & Top-2 & Top-3\\
    \hline
    GT & $0.002$ & $0.511$ & $0.703$ & $0.797$\\
    GT$\rightarrow$Joints$\rightarrow$HumanML3D &$0.015$ & $0.503$ & $0.697$ & $0.789$\\
    \cline{1-5}
    MDM~\cite{mdm}-50Step & $0.489$ & $0.455$ & $0.645$ & $0.749$\\
    MDM~\cite{mdm}-50Step-Reproduce & $0.481$  & $0.459$ & $0.651$ & $0.753$   \\
    T2M-GPT~\cite{t2m-gpt} & $0.141$ &$0.492$ &$0.679$  & $0.775$ \\
    T2M-GPT~\cite{t2m-gpt}-Reproduce & $0.115$ & $0.497$ & $0.685$ & $0.779$ \\
    MMM~\cite{mmm} & $0.089$ & $0.515$ & $0.708$& $0.804$\\
    MMM~\cite{mmm}-Reproduce & $0.071$ & $0.517$ & $0.711$& $0.805$\\
    MoMask~\cite{momask} & $\mathbf{0.045}$ & $\underline{0.521}$ & $\underline{0.713}$& $\underline{0.807}$\\
    MoMask~\cite{momask}-Reproduce & $0.093$ & $0.508$ & $0.701$ & $0.796$ \\
    \midrule
    \textbf{Ours} & $\underline{0.061}$ & $\mathbf{0.523}$ & $\mathbf{0.715}$& $\mathbf{0.810}$\\
  \end{tabular}
  }
    \vspace{-0.3cm}
\end{table}
\begin{table}[t]
\renewcommand{\arraystretch}{1.2}
\caption{\textbf{Model Scaling} results of our model. Increasing model size results in better overall performance on HumanML3D.}
\vspace{-1em}
\label{tab:scaling}
\begin{center}
\resizebox{\linewidth}{!}{%
\begin{tabular}{c|c|c|c|c|c|c}
\toprule
\multirow{2}{*}{Size} & \multirow{2}{*}{Transformer} & \multirow{2}{*}{MLP} & \multirow{2}{*}{FID $\downarrow$} & \multicolumn{3}{c}{R-Precision $\uparrow$} \\
\cmidrule(r){5-7}
 & & &  & Top 1 & Top 2 & Top 3 \\
\midrule
S & \multicolumn{1}{p{2.2cm}|}{\centering 6 head \\ 384 dim} & \multicolumn{1}{p{2.2cm}|}{\centering 3 layers \\ 1024 dim} & $0.278$ & $0.481$ & $0.676$ & $0.779$ \\
\midrule
\multirow{2}{*}{M}& \multicolumn{1}{p{2.2cm}|}{\centering 6 head \\ 384 dim} & \multicolumn{1}{p{2.2cm}|}{\centering 8 layers \\ 1280 dim} & $0.189$ & $0.479$ & $0.676$ & $0.779$ \\
\cmidrule(r){2-7}
& \multicolumn{1}{p{2.2cm}|}{\centering 12 head \\ 768 dim} & \multicolumn{1}{p{2.2cm}|}{\centering 8 layers \\ 1280 dim} & $0.173$ & $0.477$ & $0.679$ & $0.780$ \\
\midrule
\multirow{2}{*}{L} & \multicolumn{1}{p{2.2cm}|}{\centering 6 head \\ 384 dim} & \multicolumn{1}{p{2.2cm}|}{\centering 12 layers \\ 1536 dim} & $0.137$ & $0.485$ & $0.683$ & $0.785$ \\
\cmidrule(r){2-7}
& \multicolumn{1}{p{2.2cm}|}{\centering 12 head \\ 768 dim} & \multicolumn{1}{p{2.2cm}|}{\centering 12 layers \\ 1536 dim} & $0.125$ & $0.487$ & $0.685$ & $0.785$ \\
\midrule
XL &\multicolumn{1}{p{2.2cm}|}{\centering 16 head \\ 1024 dim} & \multicolumn{1}{p{2.2cm}|}{\centering 16 layers \\ 1792 dim} & $0.116$ & $0.492$ & $0.690$ & $0.790$ \\
\bottomrule
\end{tabular}}
\vspace{-2em}
\end{center}
\end{table}

\begin{figure*}[tb]
\centering
\includegraphics[width=0.9\linewidth]{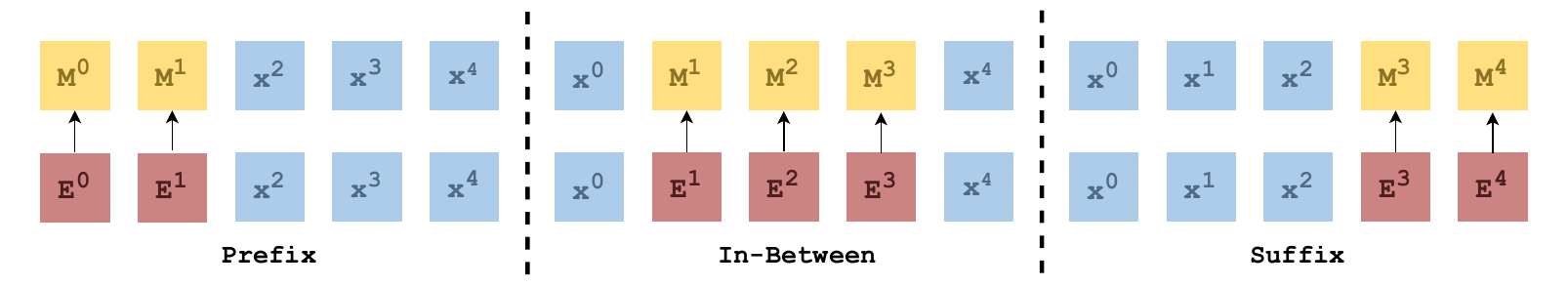}
\vspace{-1em}
\caption{\textbf{Our Method's Temporal Editing} process, including prefix, in-between, and suffix editing. The editing latents (red color) are treated as masked latents (yellow color). The sequence is then input into the generation branch in \cref{fig:architecture} to generate edited latents conditioned on the editing textual instruction and non-edit latents (blue color).}
\vspace{-0.5em}
\label{fig:editing}
\vspace{-0.3em}
\end{figure*}
\begin{table*}[t]
\renewcommand{\arraystretch}{0.6}
\caption{\textbf{Average Inference Time Results Comparison} between our method and baseline methods.}
\vspace{-1.5em}
\label{tab:speed}
\begin{center}
\resizebox{0.75\linewidth}{!}{%
\begin{tabular}{c|c|c|c|c|c|c|c}
\toprule
Methods & MDM~\cite{mdm}& MotionDiffuse~\cite{motiondiffuse}& T2M-GPT~\cite{t2m-gpt} & MLD~\cite{mld} & MMM~\cite{mmm} & MoMask~\cite{momask}& Ours\\
\midrule
AIT & 14.31s& 7.35s& 0.32s& 0.21s& 0.06s& 0.04s& 2.4s\\
\bottomrule
\end{tabular}}
\vspace{-2.5em}
\end{center}
\end{table*}

\section{Implementation Details}
\label{app:implementation_details}
For our method, the AutoEncoder is a 3-layer ResNet-based encoder-decoder with a hidden dimension of 512 and a total downsampling rate of 4.
For the generation branch, we utilize a single-layer AdaLN-Zero transformer encoder with a hidden dimension of 1024 and 16 heads as our masked autoregressive transformer. The diffusion MLPs consist of 16 layers with a hidden dimension of 1792. We also present the model scalability results in \cref{app:scaling}.

During training, we use the AdamW optimizer with $\beta_1 = 0.9$ and $\beta_2 = 0.99$. Following prior works \cite{humanml3d, t2m-gpt, mmm, momask}, the batch size is set to 256 and 512 for training the AutoEncoder on the HumanML3D and KIT-ML datasets, respectively, with each sample containing 64 frames. For training the generation branch, the batch size is set to 64 for HumanML3D and 16 for KIT-ML, with a maximum sequence length of 196 frames.
The learning rate is set at $2 \times 10^{-4}$ with a linear warmup of 2000 steps. We train the AutoEncoder for 50 epochs and modify the learning rate to decay by a factor of 20 or 10 at milestones of 150,000 and 250,000 iterations for HumanML3D and KIT-ML datasets, respectively. For the generation branch, the learning rate decays by a factor of 0.1 at 50,000 iterations for HumanML3D and 20,000 iterations for KIT-ML during a 500-epoch training process. Following image diffusion works~\cite{dit, sit, ldm}, we also incorporate exponential moving average (EMA) when updating the model parameters to achieve more stable performance.
In the generation process, for HumanML3D, the CFG~\cite{cfg} scale is set to 4.5 and for KIT, the conditioning scale is set to 2.5.

\section{Additional Quantitative Results}
\label{app:additional_quantitative}
\subsection{AutoEncoder Reconstruction Results}
In \cref{tab:ae}, we present the reconstruction results of VQ-VAE from T2M-GPT~\cite{t2m-gpt}, RVQ-VAE from MoMask~\cite{momask}, VAE from MLD~\cite{mld}, and the AutoEncoder (AE) in our method. Our AutoEncoder has much better reconstruction capability than baseline methods, which ultimately benefits both diffusion model training and sampling.
\subsection{Baseline Methods Training With Reformed Data Representation and Distribution}
In \cref{tab:mdm_new}, we demonstrate that training baseline methods using only essential dimensions can already lead to significant improvements, and processing into latent space may further enhance results.
\subsection{Original Evaluation Results}
The original evaluator is flawed due to the unnecessary focus on redundant motion representations and the new evaluators are proposed to deal with this issue. Therefore, we strongly discourage utilizing the original evaluation method to access all methods.
Also, using the original evaluator requires additional processing to convert our outputs to joints and back to the redundant representations. This inevitably introduces errors, and loses one motion frame (from joints to HumanML3D representations), and thus unfairly penalizes our method. Nevertheless, we include results in \cref{tab:original_evaluator}. Note, our method is converted and thus may be biased and inaccurately evaluated under this unfair circumstance. And notably, our reproduced MoMask exhibits worse results, similar to issues reported in their GitHub (Issues 27, 43, 99), and even ground truth motions were penalized due to the additional operations.
\subsection{Further Ablation Study and Optimize Routine}
In \cref{tab:optmize_routine}, we provide a further ablation study and an optimization routine starting from an MDM-based cosine schedule, $\boldsymbol{\epsilon}$ prediction approach to our approach. The results demonstrate the advantage of masked regression over original diffusion and the importance of our further optimization (motion representation reformation) over pure adoption of image MAR.
\subsection{Model Scalability}
\label{app:scaling}
We train six versions of our proposed model (DDPM approach), varying three transformer sizes and four diffusion MLP sizes (S, M, L, XL). These models range in size from around 30M, 100M, 180M, to 290M parameters. The performance results are summarized in \cref{tab:scaling}. We observe that increasing the model size, particularly the diffusion MLPs size, improves overall generation performance, especially in terms of FID.

\section{Temporal Editing}
\label{app:editing_inpainting}
Our method is capable of performing temporal editing in a zero-shot manner (\ie utilizing the model trained for text-to-motion generation without any editing-specific fine-tuning). In our method, temporal motion editing is easily achieved by treating the latents that need to be edited as masked latents and then generating motions following our standard generation procedure in \cref{sec:method2} which is conditions on the unmasked tokens (\ie non-edit latents) and the editing textual instructions. We visually illustrate this process in \cref{fig:editing} and we also include temporal editing results in the locally-run, anonymous HTML file referenced in \cref{app:additional_qualitative}.

\section{Additional Qualitative Results}
\label{app:additional_qualitative}
Beyond the qualitative results presented in the main paper, we also provide comprehensive video visualizations hosted on a locally-run, anonymous HTML webpage to further demonstrate the effectiveness of our approach. These visualizations include additional comparisons with state-of-the-art baseline methods, showcasing that our method generates more realistic motions and adheres more closely to textual instructions. We also present motion videos from our ablation studies to highlight the significance of each component. For example, omitting motion representation reformulation results in noticeable shaking and poses inaccuracies, while excluding the autoregressive modeling approach leads to worse textual instructions following. Furthermore, we also demonstrate our method's capability for temporal editing with prefix, in-between, and suffix editing results. Finally, we provide additional visualizations to illustrate that our method can generate a wide range of diverse and contextually appropriate motions.

\section{Limitations}
\label{app:limitation}
Since our method incorporates both standard reverse-diffusion processes (over $T$ time steps) and autoregressive generation within each step to produce high-quality and diverse motion, it inherently requires more time for motion generation compared to some baseline methods (\eg, MoMask, MMM).
To provide a clear comparison, in \cref{tab:speed}, we report the efficiency of motion generation in terms of average inference time (AIT) over 100 samples on a single Nvidia 4090 device. Notably, our method still outperforms several diffusion-based methods, \eg MDM and MotionDiffuse, in generation speed by a significant margin. For future work, we aim to explore strategies to optimize and accelerate both standard reverse-diffusion and autoregressive generation processes.

\end{document}